\documentclass[11pt]{article}

\usepackage[preprint]{acl}

\usepackage{times}
\usepackage{latexsym}
\usepackage[T1]{fontenc}
\usepackage[utf8]{inputenc}
\usepackage{microtype}
\usepackage{inconsolata}
\usepackage{graphicx}
\usepackage{booktabs}
\usepackage{xcolor}
\usepackage{amsmath}
\usepackage{amssymb}
\usepackage{placeins}
\usepackage{tabularx}

\definecolor{SimInterBase}{HTML}{105F82}
\definecolor{SimInterNoEnrich}{HTML}{497DCE}

\title{SimInter: Evidence-Grounded User Simulation with Interview-Based Profile Enrichment for Conversational Recommender Systems}

\author{
  \textbf{Erfan Shahabi\textsuperscript{*1}, Amirhossein Roshandel\textsuperscript{*1}, Mohammad Sadegh Sadeghi\textsuperscript{1}, and Azadeh Shakery\textsuperscript{1,2}} \\
  \textsuperscript{1}School of Electrical and Computer Engineering, College of Engineering, University of Tehran, Tehran, Iran \\
  \textsuperscript{2}Institute for Research in Fundamental Sciences (IPM), Tehran, Iran \\
  \texttt{\{erfanshahabi, roshandel.amir, mhmssadeghi74, shakery\}@ut.ac.ir}
}

\begin{document}
\maketitle
\renewcommand{\thefootnote}{*}
\footnotetext{Equal contribution, ordered randomly.}
\renewcommand{\thefootnote}{\arabic{footnote}}

\begin{abstract}
Conversational Recommender Systems (CRSs) require reliable user
simulators for cost-effective evaluation. LLM-based simulators
have made realistic simulation possible but typically condition on
profiles derived from explicit ratings, demographics, and static
sources capturing surface preferences while omitting implicit
behavioral characteristics such as decision-making style,
aspect-level preferences, and domain expertise. We propose
\textit{SimInter}, a pipeline that elicits these attributes via
synthetic interviews between two LLM agents, grounded in the user's
statistical profile and historical conversational evidence. Before
profile enrichment, an audit agent explicitly flags potential
fabrication and stereotyping at the turn level. We instantiate and
evaluate SimInter on movie recommendation. We simulate $300$ ReDial
users and, on a matched subset of $100$ users, compare SimInter with a
statistics-only baseline: SimInter significantly increases the
diversity of expressed preference aspects (filtered aspect entropy:
$1.619$ vs.\ $0.859$; paired $d=0.96$, Holm-adjusted $p<0.001$). This
gain persists when the simulator converses with two trained
recommenders, KBRD and UniCRS, under two independent evaluation
judges ($+0.28$ to $+0.56$ nats; all $p<0.001$). Human evaluation finds $94.3\%$ of enriched-attribute
claims supported or partially supported, while
human judgments agree with $95.2\%$ of the auditor's
binary filtering decisions on a stratified sample.
\end{abstract}
\section{Introduction}

Conversational Recommender Systems (CRSs) have become a primary 
interface for content 
discovery~\cite{Fu2020TutorialOC, Kim2012CustomerDrivenCR}, but their 
evaluation remains difficult: studies with real users are expensive, 
slow, and rarely reproducible, with outcomes depending on the specific 
sample recruited~\cite{10.1145/3589335.3651955}. User simulation has 
emerged as a complementary methodology in which artificial users 
interact with a CRS in place of humans, enabling controlled, repeatable 
experimentation at scale~\cite{Balog2025UserSI, Zhu2024ALC}.

Early simulators were rule-based or agenda-based~\cite{Ru2012RuleBasedAF, 
Afzali_2023}, offering controllability at the cost of behavioral 
realism. Large language models (LLMs) have changed this picture, but 
LLM-based simulators still exhibit characteristic failure modes: 
repetitive responses, narrow choice ranges, binary reactions, and drift 
toward generic ``average user'' 
behavior~\cite{yoon2024evaluating, 10.1145/3589335.3651955}. We argue
that an important source of these limitations is the user profile that
conditions the simulator.

\paragraph{The profile bottleneck.}
Most LLM-based simulators condition generation on profiles built from
observable signals: ratings, interaction histories, and demographic
attributes. Such profiles capture \emph{what} a user has consumed but
leave largely unspecified \emph{how} the user makes choices: whether
they satisfice or maximize, whether their choices are driven by mood
or social context, and how willing they are to accept unfamiliar
recommendations. These behavioral dispositions strongly shape
conversational behavior yet are precisely what ratings-based profiles 
omit~\cite{tigunova2019listening, yoon2024evaluating}. A natural
alternative---prompting an LLM to infer behavioral attributes from a
static profile---has a known failure mode: LLMs tend to produce 
stereotyped inferences, inflating apparent profile richness while 
reducing individual fidelity. The same risk applies to self-play 
dialogue, where two agents can amplify rather than reveal latent 
attributes, fabricating plausible-sounding traits not grounded in 
actual behavior.

\paragraph{Our approach.}
We propose \textit{SimInter}, which derives behavioral attributes from
synthetic dialogue while explicitly defending against fabrication via
four design choices.
\emph{(i) Statistical grounding}: we first compute behavioral statistics
(acceptance rate, reference style, genre distribution, exploration
ratio) from the user's interaction history; the synthetic dialogue
is then conditioned on these statistics and the user's historical
conversational evidence.
\emph{(ii) Interview rather than peer dialogue}: generation is cast as 
an asymmetric interview, with the interviewee anchored to the user's 
profile and the interviewer as a neutral elicitor probing specific 
dimensions~\cite{tigunova2019listening}, preventing mutual 
reinforcement of speculative content.
\emph{(iii) Turn-level auditing}: an \textit{Audit Agent}
flags fabrication, history contradictions, and caricature
(generic persona templates and invented rationalizations).
Depending on severity, flagged turns are excluded from
profile enrichment, while structurally compromised
dialogues trigger regeneration.
\emph{(iv) Evidence-linked enrichment}: an LLM agent enriches the
user's statistical profile with attributes extracted from the audited
interview, with each categorical or scored attribute paired with a
supporting evidence quote; human annotators verify that $94\%$ of these
enriched claims are supported by the interview.
\paragraph{Contributions.}
(1)~We identify the \emph{profile bottleneck} in LLM-based user
simulation. (2)~We propose \textit{SimInter}, combining statistical
grounding, asymmetric interview dialogue, turn-level auditing, and
evidence-linked profile enrichment; the audit uses a flag taxonomy
that distinguishes unsupported claims from invented rationalizations.
(3)~We evaluate \textit{SimInter} on $300$ ReDial users
($1{,}200$ simulated conversations), with matched
$100$-user comparisons against statistics-only and
CSHI~\cite{Zhu2024ALC} baselines across four
complementary tracks (item diversity, population-level
preference alignment, aspect richness, and linguistic
variety). We find a significant increase in aspect
richness over statistics-only profiles (T3, $d=+0.96$,
Holm-adjusted $p<0.001$).

\section{Related Work}

User simulation provides a scalable alternative to
human evaluation of conversational recommender
systems (CRSs)~\cite{Balog2025UserSI,Zhu2024ALC,
zhang2020evaluating,bernard2025limitations}.
Earlier approaches relied on agenda-based control,
later extended with neural generation in frameworks
such as GenTUS~\cite{Afzali_2023,
lin-etal-2022-gentus}. LLM-based frameworks,
including iEvaLM, enable more flexible interactions
but raise concerns about behavioral realism,
data leakage, and controllability
~\cite{wang-etal-2023-rethinking-evaluation,
yoon2024evaluating,10.1145/3589335.3651955}.

Recent approaches incorporate richer user models.
RecUserSim combines bounded-rationality principles
with memory tracking~\cite{Chen2025RecUserSimAR};
PerCRS uses assigned personality traits
~\cite{zhao2025exploring}, while PUB infers
Big-Five traits from behavioral logs
~\cite{ma2025pub}. PEARL derives persona knowledge
from user reviews~\cite{kim2024pearl}, and CSHI
offers controllable, target-conditioned simulation
~\cite{Zhu2024ALC}. Other work investigates
LLM-based and user-centric CRS evaluation and
parameterized behavioral traits
~\cite{Chen2025EvaluatingCR,10.1145/3631534,
10.1145/3673791.3698425}.

SimInter differs by eliciting implicit behavioral
attributes through synthetic interviews grounded
in historical conversational evidence. Unlike
predefined trait configurations or review-derived
personas, SimInter uses user-specific dialogue
and audits potentially fabricated or stereotyped
content before profile enrichment.

Table~\ref{tab:related_comparison} compares these
frameworks in terms of datasets, profile sources,
available ground truth, target conditioning,
reliability mechanisms, evaluation objectives,
and data requirements. Notably, while CSHI
addresses target-item leakage and PerCRS uses
knowledge-graph grounding, SimInter audits
interview-derived claims against historical
user evidence before profile enrichment.

\begin{figure*}[t]
  \centering
  \includegraphics[width=\textwidth]{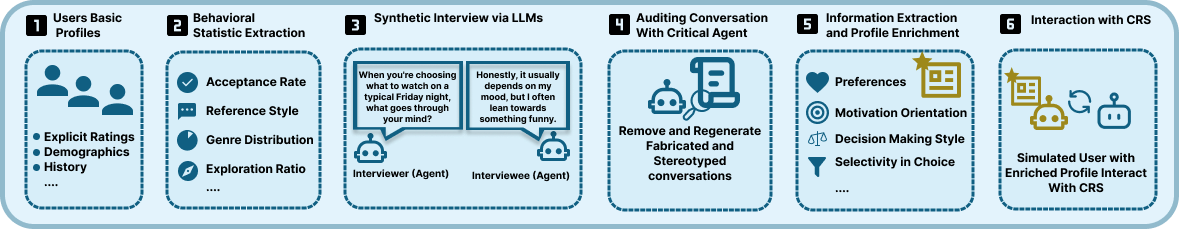}
  \caption{Overview of the SimInter pipeline.}
  \label{fig:approach}
\end{figure*}

\begin{table*}[t]
\centering
\footnotesize
\setlength{\tabcolsep}{4pt}
\renewcommand{\arraystretch}{1.12}

\begin{tabularx}{\textwidth}{
@{}>{\raggedright\arraybackslash}p{2.3cm}
*{4}{>{\raggedright\arraybackslash}X}@{}}
\toprule
\textbf{Dimension} &
\textbf{CSHI}~\cite{Zhu2024ALC} &
\textbf{PerCRS}~\cite{zhao2025exploring} &
\textbf{PUB}~\cite{ma2025pub} &
\textbf{SimInter (ours)} \\
\midrule

Dataset
& ReDial, OpenDialKG, MovieLens
& DuRecDial 2.0
& Amazon Reviews
& ReDial \\

Conversational
& Yes & Yes & No & Yes \\

Profile source
& History; generated or editable profiles
& History; injected Big Five
& Behavioral logs; inferred Big Five
& Statistics; audited interviews \\

Profile ground truth
& Target items; held-out interactions
& Injected personality traits
& Held-out real interactions
& Historical preferences; no trait labels \\

Target-item conditioning
& Yes & Yes & N/A & No \\

Fabrication control
& Target-leakage prevention
& KG-based factual grounding
& Not reported
& Evidence-based turn-level auditing \\

Evaluation target
& CRS recommendation performance
& Trait consistency; CRS outcomes
& Behavioral fidelity; RS ranking
& Simulator behavior; human evaluation \\

Data requirements
& Interaction history; item information
& Preferences; traits; KG
& Ratings; reviews; timestamps; metadata
& Dialogues; historical preferences \\

\bottomrule
\end{tabularx}

\caption{Conceptual comparison of user simulation
frameworks. Ground truth denotes available evaluation
references, not necessarily verified personality traits.
Fabrication-control mechanisms differ in scope;
N/A indicates not applicable.}
\label{tab:related_comparison}
\end{table*}

\section{Method}
\label{sec:method}

Figure~\ref{fig:approach} illustrates the \textit{SimInter} pipeline.
The method consists of six stages:
(1)~constructing basic user profiles from historical interactions,
(2)~extracting behavioral statistics from these profiles,
(3)~generating synthetic interviews between two LLM-based agents,
(4)~auditing the generated conversations with an audit agent,
(5)~extracting implicit user attributes and enriching the profiles,
and (6)~simulating users interacting with a CRS using their
enriched profiles. We instantiate the LLM-based components with
\textit{Gemini 2.5 Flash}~\cite{comanici2025gemini}
for interview generation and user simulation,
\textit{Claude Haiku 4.5}~\cite{anthropic2025haiku}
for auditing, \textit{Tencent Hy3}~\cite{tencent2026hy3} for enrichment,
and \textit{GPT-4o-mini}~\cite{achiam2023gpt}
as the CRS the simulated user interacts with.
Models were chosen for availability and cost; where possible, we drew
components from different model families to reduce shared bias between
generator, auditor, and judges. All model choices are flexible.

\subsection{Basic User Profiles and Behavioral Statistics}
\label{sec:s1}

\textbf{Stage 1 (Basic User Profiles).}
We construct basic user profiles from
ReDial~\cite{li2018towards}, treating each crowdworker as a user across
both seeker and recommender roles. Of the 1{,}112 workers in the
$11{,}348$ ReDial dialogues, we retain the 618 with at least two
conversations. Each profile contains observed movie preferences
(liked, disliked, and implicitly disliked movies, derived from
ReDial's post-conversation questionnaires) and the user's historical
conversations.

\begin{figure}[t]
  \centering
  \includegraphics[width=\columnwidth]{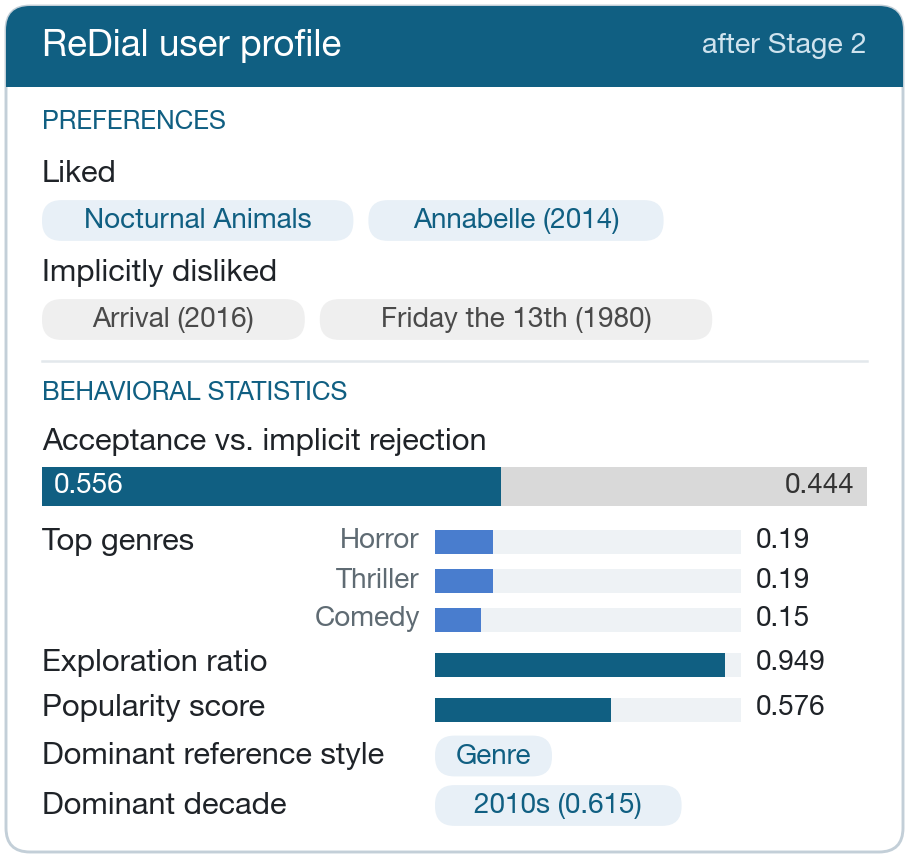}
  \caption{Example ReDial user profile after Stage 2,
  combining historical preferences with derived behavioral statistics.}
  \label{fig:profile_example}
\end{figure}

\noindent\textbf{Stage 2 (Behavioral Statistic Extraction).}
We augment these profiles with rule-based behavioral
statistics derived from historical interactions: acceptance and rejection rates, 
reference style (genre, mood, actor, director, year), genre and decade 
distributions, exploration ratio, and popularity score. Together these 
form a behavioral fingerprint that grounds the synthetic interview. 
Figure~\ref{fig:profile_example} shows an example. Of the 618 profiles,
543 meet our eligibility criteria, requiring at least one seeker-role
conversation and a nonzero acceptance or implicit-rejection signal.
We randomly sample 300 users from this eligible set for the full
SimInter pipeline, and randomly select a subset of 100 of these users
for matched ablation experiments and comparison with CSHI
(\S\ref{sec:eval-setup}).

\subsection{Synthetic Interview Generation}
\label{sec:s2}
\noindent\textbf{Stage 3 (Synthetic Interview Generation).}

Conversations often contain implicit information that provides deeper
insights into
users~\cite{tigunova2019listening, torbati2021you}. We exploit LLMs'
role-playing capabilities~\cite{wang-etal-2024-rolellm, chen2024personapersonalizationsurveyroleplaying}
to simulate interviews between two agents: an interviewee anchored to
the user's profile and an interviewer that elicits it. For each of the
300 sampled users, the interviewee agent receives the user's
statistical profile and one historical ReDial conversation in which
the user acts as a seeker. The interviewer, in contrast, has no access
to the profile or the ReDial history; it sees only the ongoing
interview, so that it cannot steer the interviewee toward known
preferences. It opens with one of four predefined open-ended
questions, chosen at random (e.g., \textit{``Walk me through the last
movie you really enjoyed --- what was it and what stuck with you?''};
all four are listed in Appendix~\ref{sec:appendix-prompts-interviewer}), and
dynamically adapts its follow-up questions to the interviewee's
responses. Each interview comprises an initial six-turn elicitation
phase followed by four clarification turns, for a total of 10 turns.
Figure~\ref{fig:conversation} shows an example; full system prompts
are in Appendix~\ref{sec:appendix-prompts}.

\begin{figure}[t]
  \centering
  \includegraphics[width=\columnwidth]{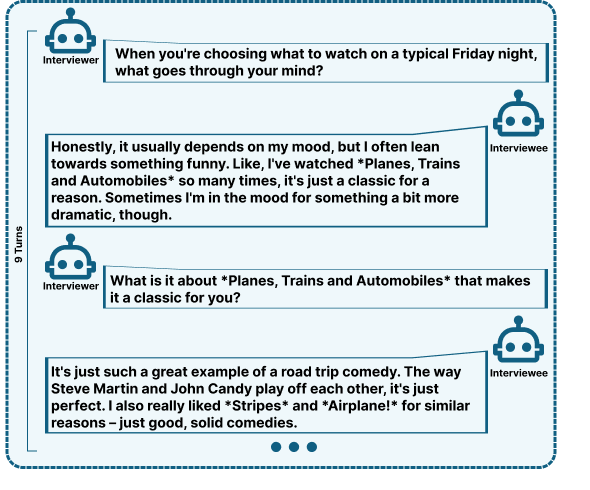}
  \caption{Example simulated interview between interviewer and 
  interviewee agents.}
  \label{fig:conversation}
\end{figure}

\subsection{Conversation Auditing}
\label{sec:s3}
\noindent\textbf{Stage 4 (Conversation Auditing).}

An \textit{Audit Agent} checks each interviewee turn against the
user's own evidence, namely the statistical profile and the historical
ReDial conversation, for three types of issues.
\emph{Fabrication} covers specific claims that are neither stated in
nor plausibly inferable from this evidence, such as invented viewing
experiences or personal anecdotes.
\emph{Caricature} covers stereotyped persona performances and
\emph{invented rationalizations}: plausible-sounding justifications
for preferences that the profile does not attest, or recognizable
persona archetypes it does not support.
\emph{History contradiction} covers statements that conflict with the
user's liked and disliked movies or with their earlier ReDial
conversation. In addition, deterministic structural checks verify the
number of turns, the alternation of speaker roles, and the absence of
empty turns.

The auditor assigns each flag a severity (low, medium, or high), and a
deterministic rule derives the dialogue-level decision. Any structural
error or $\geq 3$ high-severity flags trigger \textsc{regenerate}
(up to two retries, with a corrective nudge appended to the
interviewee prompt). At least one high-severity or $\geq 3$
medium-severity flags trigger \textsc{partial}: the interview is
retained, but its high- and medium-severity turns are excluded from
enrichment. All other interviews are accepted (\textsc{accept}).
For example, in one interview, the interviewee claimed to have enjoyed
\textit{Blue Streak}, although the user's ReDial conversation
indicated that they had not seen it. In another, the interviewee
dismissed the action in \textit{Justice League} as hard to care about,
although the film is on the user's liked list. The auditor flagged
both turns, so they could be excluded before profile enrichment. Full
auditor prompts, flag definitions, and representative examples are in
Appendix~\ref{sec:appendix-auditor}.

\subsection{Profile Enrichment}
\label{sec:s4}
\noindent\textbf{Stage 5 (Profile Enrichment).}

From audited interviews, we extract six implicit user attributes 
grounded in prior user-modeling 
literature: \emph{aspect-level 
preferences}~\cite{radlinski-etal-2019-coached}, 
\emph{pickiness}~\cite{yoon2024evaluating}, 
\emph{decision-making style}~\cite{Scott1995DecisionMakingST, hayati-etal-2020-inspired}, 
\emph{viewing motivation}~\cite{oliver2011entertainment}, 
\emph{novelty preference}~\cite{tang2022preference}, and 
\emph{domain expertise}~\cite{kostric2025know}. Each attribute reveals 
behavior not observable in the static profile: aspect preferences 
organize preferences along Movie/Genre/Person/Other anchors; pickiness 
captures selectivity; decision-making style ranges across rational, 
intuitive, dependent, avoidant, and spontaneous; viewing motivation 
combines a hedonic-eudaimonic score (1--5) with a primary category 
(escapism, hedonic, eudaimonic, social, catharsis, arousal); novelty 
preference scores openness to unfamiliar content (1--5); and domain 
expertise classifies users as novice, intermediate, or cinephile. 
Full attribute schemas, evidence-extraction prompts, and an example 
enriched profile are in Appendix~\ref{sec:appendix-enrich}.

The inferred attributes and derived pickiness are combined
with the Stage~2 statistical profile to form the enriched
profile that conditions user simulation in Stage~6.

\subsection{User Simulation}
\label{sec:s5}
\noindent\textbf{Stage 6 (User Simulation).}

The enriched profile of each user conditions 
\textit{Gemini 2.5 Flash}~\cite{comanici2025gemini} to simulate that 
user's behavior in conversation with a CRS powered by 
GPT-4o-mini~\cite{achiam2023gpt}. Each user is simulated in four 
conversations varying the opening strategy across \emph{aspect}, 
\emph{movie}, \emph{mood}, and \emph{expertise} prompts. After each
CRS turn, starting from the fourth, an LLM termination judge
classifies the last exchange as \emph{true acceptance} (the user
explicitly commits to a specific recommendation), \emph{resigned
acceptance}, \emph{explicit give-up}, \emph{other}, or
\emph{continue}; any label other than \emph{continue} ends the
conversation, which is otherwise capped at eight CRS turns.
\section{Evaluation}
\label{sec:evaluation}

We evaluate SimInter on ReDial using 300 users in the
full pipeline and a matched subset of 100 users for
comparisons against two baselines:
\textsc{StatisticalOnly} and CSHI.
We ask four questions: do enriched 
users (T1)~mention a more diverse set of items? (T2)~express preferences correlated with population-level movie ratings? (T3)~discuss movies 
along richer behavioral aspects? (T4)~differ in surface linguistic 
variety? We additionally conduct a targeted human evaluation
(\S\ref{sec:human-eval}) and report a separate cross-CRS
robustness experiment with two trained ReDial recommenders, KBRD~\cite{chen2019knowledgebasedrecommenderdialog} and
UniCRS~\cite{Wang_2022} (Appendix~\ref{sec:appendix-cross-crs}). Our evaluation focuses on
\textit{profile-conditioned simulated user behavior}, rather than the
recommendation accuracy of the underlying CRS models.

\subsection{Experimental Setup}
\label{sec:eval-setup}

\textsc{Full} represents the complete SimInter pipeline,
combining behavioral statistics, audited interviews,
and enriched user attributes.
\textsc{StatisticalOnly} uses only the behavioral
statistics, without interview-based profile enrichment.
We additionally compare against CSHI, an existing
user-simulation baseline that follows a target-conditioned
dialogue protocol.

We run \textsc{Full} on 300 ReDial users, generating
four conversations per user (1,200 conversations).
For matched comparisons, we select a subset of 100
users and evaluate \textsc{StatisticalOnly} and CSHI
on the same users, generating 400 conversations
per baseline.
All comparisons use the corresponding 100-user
subset of \textsc{Full}.

We report descriptive metrics for the full 300-user
cohort and conduct paired comparisons on the matched
100-user subset. For each comparison, we compute
user-level mean differences with bootstrap 95\%
confidence intervals (20,000 paired resamples),
two-sided paired sign-flip tests (50,000 randomizations),
and paired Cohen's $d_z$~\cite{Cohen2013-ty}.
We apply Holm's correction~\cite{29def780-e117-38f0-8afb-edf384af3fad}
across the original family of 24 paired metric tests,
using a significance level of $\alpha=0.05$.
T2 population-level correlations are reported
descriptively.

\subsection{Evaluation Tracks}
\label{sec:eval-tracks}

\paragraph{T1 — Item diversity.} For each user, Shannon entropy of the 
movie-mention distribution across their four conversations. Higher 
entropy indicates broader coverage, reflecting whether the simulator 
escapes the ``average user'' attractor~\cite{10.1145/3589335.3651955}. 
We also report Jensen--Shannon divergence (JSD) against the movie-mention distribution of ReDial seekers.

\paragraph{T2 — Popularity alignment.} For movies the user rates or 
reacts to in conversation, we correlate the simulator's expressed 
preference with the movie's mean MovieLens-25M 
rating~\cite{harper2015movielens} (Pearson and Spearman). This tests 
whether stated preferences track population-level signal.

\paragraph{T3 — Aspect richness.} Using Claude Haiku
4.5~\cite{anthropic2025haiku} as an aspect-level
judge~\cite{Liu2023-sg, Zheng2023-ft}, we extract
\textit{(aspect, sentiment)} pairs from the user's sentences about
each movie discussed in the conversation. After filtering generic
placeholders (e.g., \textit{movie}, \textit{film}, \textit{story}),
we compute per-user Shannon entropy over remaining aspects and over
the sentiment distribution. Higher aspect entropy indicates the
simulator articulates preferences along a wider range of behavioral
dimensions rather than collapsing to a single template.

\paragraph{T4 — Linguistic surface variety.} Per-user metrics over
user turns: corpus-level type-token ratio (TTR), n-gram monotony
(mean Jaccard overlap of bigrams/trigrams between turn pairs), and
SBERT semantic diversity~\cite{Reimers2019-ej}. TTR and n-gram
monotony measure \emph{lexical} variety, SBERT diversity measures the
\emph{semantic} spread of requests, and T3 captures
\emph{semantic-content} variety; these can move oppositely.
\subsection{Results}
\label{sec:eval-results}
Table~\ref{tab:headline} summarizes the matched
100-user comparisons. Relative to
\textsc{StatisticalOnly}, \textsc{Full} achieves
higher T3 aspect entropy ($\Delta=+0.760$)
but lower T1 item diversity ($\Delta=-0.375$),
both significant after Holm correction.
\textsc{Full} also exceeds CSHI on T1, T3, and T4.
T2 correlations are reported descriptively.
\begin{table*}[t]
\centering
\small
\setlength{\tabcolsep}{5pt}
\begin{tabular}{llrrrrr}
\toprule
\textbf{Track} & \textbf{Metric}
& \textbf{\textsc{Full}}
& \textbf{\textsc{StatOnly}}
& \textbf{CSHI}
& \textbf{$\Delta_{F-S}$}
& \textbf{$\Delta_{F-C}$} \\
\midrule
T1 & Item entropy
& 2.034 & 2.409 & 1.454
& $-0.375^{\star}$ & $+0.588^{\star}$ \\
T1 & JSD vs.\ ReDial
& 0.681 & 0.676 & 0.682
& $+0.005^{\star}$ & $-0.001$ \\
\midrule
T2 & Population Pearson $r$
& 0.171 & 0.133 & 0.074
& --- & --- \\
\midrule
T3 & Aspect entropy (filtered)
& 1.619 & 0.859 & 0.194
& $+0.760^{\star}$ & $+1.425^{\star}$ \\
T3 & Aspect entropy (unfiltered)
& 1.857 & 1.490 & 0.694
& $+0.366^{\star}$ & $+1.163^{\star}$ \\
\midrule
T4 & SBERT diversity (pairwise)
& 0.532 & 0.500 & 0.345
& $+0.032$ & $+0.186^{\star}$ \\
T4 & SBERT diversity (centroid)
& 0.226 & 0.211 & 0.140
& $+0.016$ & $+0.086^{\star}$ \\
\bottomrule
\end{tabular}
\caption{Evaluation results on the matched 100-user
subset. \textsc{StatOnly} denotes \textsc{StatisticalOnly}.
$\Delta_{F-S}$ and $\Delta_{F-C}$ denote paired
mean differences between \textsc{Full} and
\textsc{StatisticalOnly} and CSHI, respectively.
$\star$ indicates significance after Holm correction
($p_{\mathrm{adj}}<0.05$). CSHI T1 is defined for
94 users (six mention a single movie).
T2 correlations are descriptive and computed
over condition-specific movie sets.}
\label{tab:headline}
\end{table*}

\paragraph{T1 — Item diversity.}
\textsc{Full} exhibits a more concentrated movie-mention distribution
than \textsc{StatisticalOnly}
($p_{\mathrm{adj}}<0.001$), and a marginally higher but significant JSD
from ReDial ($0.681$ vs.\ $0.676$). These results characterize the
breadth of movie mentions rather than the fidelity of simulated
preferences.

\paragraph{T2 — Popularity alignment.}
In the full 300-user evaluation, SimInter exhibits
a moderate positive correlation between simulated
movie preferences and MovieLens-25M mean ratings
(Pearson $r=0.347$, $n=275$ movies). This suggests
that the preferences expressed in simulated dialogue
capture meaningful population-level rating patterns.
On the matched 100-user subset, correlations remain
positive for \textsc{Full} ($r=0.171$) and
\textsc{StatisticalOnly} ($r=0.133$).
As the movie sets differ across conditions, these
estimates are descriptive rather than paired
comparisons and do not establish individual-level
preference fidelity. Further diagnostics are provided
in Appendix~\ref{sec:appendix-t2}.

\paragraph{T3 — Aspect richness.}
One of the strongest findings is the substantial
increase in the diversity of movie-related aspects
expressed by enriched users. Across the full
300-user cohort, \textsc{Full} achieves a mean
aspect entropy of $1.606$. On the matched 100-user
subset, \textsc{Full} substantially exceeds
\textsc{StatisticalOnly} ($1.619$ vs.\ $0.859$;
$\Delta=+0.760$, $d_z=0.955$,
$p_{\mathrm{adj}}<0.001$).
These findings highlight the contribution of
interview-based profile enrichment to richer
aspect-level preference expression.

\begin{figure}[t]
  \centering
  \includegraphics[width=\columnwidth]{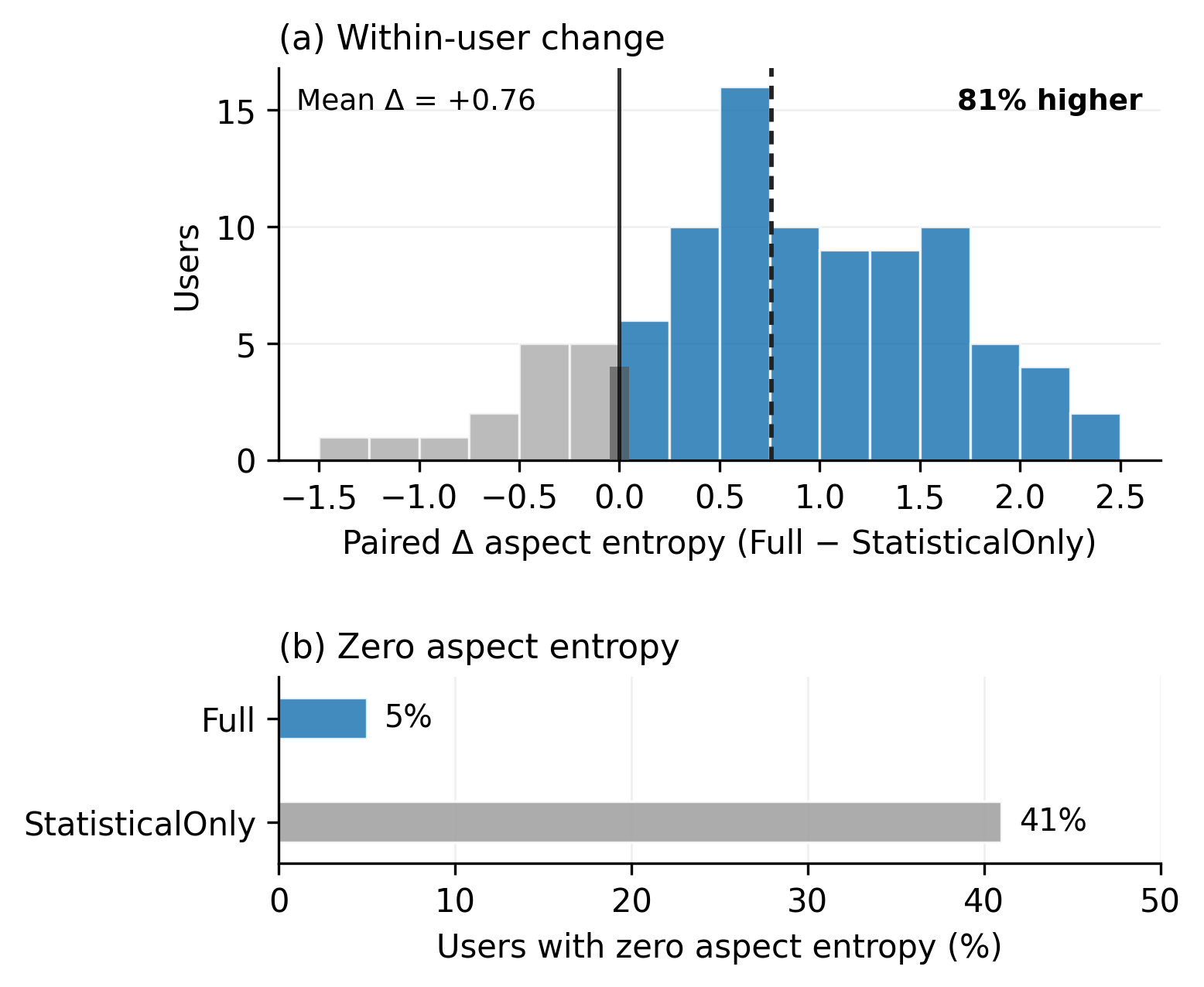}
  \caption{Effect of interview-based profile enrichment
  on filtered T3 aspect entropy across 100 matched users.
  (a) Per-user paired differences between \textsc{Full}
  and \textsc{StatisticalOnly}; the dashed line indicates
  the mean difference and the solid line indicates zero.
  (b) Percentage of users with zero aspect entropy
  under both conditions.}
  \label{fig:t3-per-user}
\end{figure}

Figure~\ref{fig:t3-per-user} shows that
\textsc{Full} achieves higher filtered aspect
entropy than \textsc{StatisticalOnly} for 81
of the 100 matched users (15 lower, 4 unchanged).
Moreover, only 5\% of users under \textsc{Full}
exhibit zero aspect entropy, compared with 41\%
under \textsc{StatisticalOnly}.

A concrete example illustrates how enrichment
shapes preference expression. User 468's enriched
profile captures preferences for relentless pacing
and action choreography. In a matched
\textit{mood}-strategy opening, the \textsc{Full}
simulator requests ``an action thriller with
relentless pacing and some genuinely tight
choreography,'' whereas \textsc{StatisticalOnly}
asks for ``some intense action'' with
``a good crime or thriller plot.''
References to pacing and choreography recur
across all four \textsc{Full} openings,
illustrating how enriched profile attributes
translate into more specific movie-related
requests. Additional qualitative examples and
aspect distributions are provided in
Appendix~\ref{sec:appendix-t3}.

\paragraph{T4 — Linguistic surface variety.}
\textsc{Full} shows slightly higher SBERT semantic
diversity than \textsc{StatisticalOnly} ($0.532$
vs.\ $0.500$; $d_z=+0.25$), but the difference
is not significant ($p_{\mathrm{adj}}=0.11$).
Conversely, \textsc{Full} exhibits lower corpus
TTR ($0.674$ vs.\ $0.728$; $d_z=-0.68$) and higher
n-gram monotony ($0.028$ vs.\ $0.019$;
both uncorrected $p<0.001$).

This lexical trade-off is consistent with
the repetition of profile-specific expressions
across conversations, a known issue in
profile-conditioned generation
~\cite{10.1145/3673791.3698425}.
For example, User 468 repeatedly uses
``relentless'' across three of four conversations,
reflecting vocabulary from the enriched profile.
Thus, T3 and T4 capture complementary properties:
enrichment can increase aspect richness and semantic
diversity while narrowing lexical expression.
These distinctions matter when diagnosing CRS
breakdowns~\cite{Bernard2024IdentifyingBI},
particularly when varied user phrasing is important.

\paragraph{Comparison with CSHI.} Under its target-conditioned
protocol with the same GPT-4o-mini CRS, CSHI produces markedly less
diverse simulated behavior than \textsc{Full}: aspect entropy is
$0.194$ vs.\ $1.619$ ($d_z=1.85$), item entropy $1.454$ vs.\ $2.034$
($d_z=1.13$), and SBERT diversity $0.345$ vs.\ $0.532$ ($d_z=1.30$;
all $p_{\mathrm{adj}}<0.001$), while JSD from ReDial does not differ.
CSHI requests are also lexically repetitive (n-gram monotony $0.300$
vs.\ $0.028$), and its population correlation is weak ($r=0.074$,
$88$ movies). Because CSHI steers each conversation toward a
predefined target item, these differences reflect both profile
content and protocol, not profile enrichment alone.

\subsection{Auditor Behavior}
\label{sec:auditor-results}
The Audit Agent processed 300 interviews: \textsc{accept} ($209$,
$70\%$), \textsc{partial} ($91$, $30\%$), \textsc{regenerate} ($0$).
Only one interview reached the \textsc{regenerate} threshold; the
regenerated interview received \textsc{partial}. All other corrective
action was therefore turn-level exclusion under \textsc{partial},
removing $196$ interviewee turns from enrichment. The auditor raised
$676$ flags ($372$ caricature, $173$ history contradiction, $131$
fabrication); the full taxonomy and example dialogues are in
Appendix~\ref{sec:appendix-auditor}.

Three observations follow. First, the dominant issue is not
fabrication but caricature ($372/676$ flags, $55\%$), mostly invented
rationalizations. For example, one interviewee explained a dislike of
\textit{The Hangover} and \textit{Superbad}, both on the user's
disliked list, by calling them ``loud and obnoxious,'' although the
user likes comparably broad comedies; the dislike is attested, but
the reason is not. Second, the human evaluation shows a matching
pattern (\S\ref{sec:human-eval}): \textit{decision-making style},
which has no quantitative anchor in the profile, is the least
supported attribute ($21.4\%$ Unsupported; $28.6\%$
Evidence-grounded), whereas \textit{pickiness}, derived from the
rejection rate, is among the best supported ($97.0\%$); this
correspondence offers limited, indirect evidence of audit
calibration. Third, the absence of final \textsc{regenerate} decisions
indicates that gross structural failure is rare with capable LLMs at
our interview scale; the dominant failure mode is local, which makes
turn-level exclusion an appropriate response.

\subsection{Human Evaluation of Audit Decisions}
\label{sec:human-audit}

We evaluated Stage~4 auditing on $105$ synthetic interviewee turns,
with one turn per historical ReDial user, sampled equally from the
$35$ excluded, $35$ flagged-but-retained, and $35$ unflagged audit
outcomes. Three co-author annotators independently examined each
synthetic response alongside evidence from the corresponding user's
original ReDial conversations, without seeing the auditor's decisions,
flags, or rationales. They assigned one of three labels:
\textsc{Excluded} (a serious, evidence-supported issue warrants
removing the turn), \textsc{Flagged--Retained} (a meaningful issue
exists but does not warrant removal), or \textsc{Unflagged}
(no meaningful issue is identified). The annotators agreed
unanimously on $70/105$ turns ($66.7\%$; Fleiss' $\kappa=0.664$).

Using majority human judgments as the reference, the auditor achieved
\emph{86.7\% three-class accuracy} ($91/105$), with agreement on $31/35$
excluded, $31/35$ flagged-but-retained, and $29/35$ unflagged turns.
Collapsing both retained classes yielded \emph{95.2\% binary filtering
accuracy} ($100/105$): four auditor-excluded turns were judged
retainable and one auditor-retained turn was judged excludable.
These results suggest that Stage~4 can distinguish concerns that warrant
removal from those that need not discard an otherwise usable response.
The figures are human-referenced accuracies on this balanced sample,
not prevalence-adjusted estimates over all audited turns; annotation
protocol, disagreements, and limitations are detailed in
Appendix~\ref{sec:appendix-human-audit}.

\subsection{Human Evaluation of Enriched Attributes}
\label{sec:human-eval}

Three annotators independently evaluated enriched-
attribute claims from \textsc{Full} profiles. We report
105 claims from 45 users rated by all three annotators
(315 annotations). Using behavioral profiles and
interview transcripts, annotators assessed support,
evidence grounding, consistency, usefulness, overreach,
and confidence (Appendix~\ref{sec:appendix-human-eval}).

\paragraph{Results.}
By majority vote, 94.3\% of claims were
Supported or Partially supported (41.0\%
strictly Supported), and 5.7\% were
Unsupported or Contradicted. At the annotation
level, 1.3\% were Contradicted; mean usefulness
and overreach were 3.61/5 and 1.70/5.
Strict support was highest for pickiness
(72.7\%) and liked aspects (75.0\%) while
decision-making style had the highest
Unsupported/Contradicted rate (21.4\%;
Table~\ref{tab:human-eval}).

\paragraph{Inter-annotator agreement.}
Pairwise quadratic-weighted Cohen's $\kappa$
ranged from 0.40 to 0.52, with 97--98\%
within-one-rank agreement. For binary support,
Gwet's AC1 was 0.89, while Fleiss' $\kappa$
was lower under label imbalance.
Disagreements mainly involved Supported versus
Partially supported (Appendix~\ref{sec:appendix-human-eval}).

\begin{table}[t]
\centering
\small
\begin{tabular}{lrrr}
\toprule
\textbf{Attribute} & \textbf{$n$} & \textbf{Support} & \textbf{Grounded} \\
\midrule
Viewing motivation        & 51  & 100.0\% & 47.1\% \\
Aspect prefs.\ (liked)    & 60  & 100.0\% & 81.7\% \\
Aspect prefs.\ (descr.)   & 24  & 100.0\% & 91.7\% \\
Pickiness                 & 66  & 97.0\%  & 65.2\% \\
Domain expertise          & 36  & 88.9\%  & 41.7\% \\
Novelty preference        & 15  & 86.7\%  & 46.7\% \\
Aspect prefs.\ (disliked) & 21  & 81.0\%  & 76.2\% \\
Decision-making style     & 42  & 78.6\%  & 28.6\% \\
\midrule
\textbf{All}              & 315 & \textbf{93.3\%} & \textbf{59.7\%} \\
\bottomrule
\end{tabular}
\caption{Human-evaluation support rates by attribute.
\textit{Support} = Supported + Partially supported;
\textit{Grounded} = Evidence-grounded. $n$ is number of annotations
(each claim has three).}
\label{tab:human-eval}
\end{table}

\subsection{Cross-CRS Robustness}
\label{sec:cross-crs-robustness}

We evaluated T3 robustness on 120 users with KBRD
and UniCRS under two independent judges. After
audit filtering, the paired samples contained 111
KBRD and 80 UniCRS users. The aspect-entropy
improvement persisted for KBRD ($\Delta=+0.265$
and $+0.280$) and UniCRS ($\Delta=+0.330$ and
$+0.369$) under GPT-4o-mini and DeepSeek-chat,
respectively (all $p<0.001$). Including all
technically completed conversations also yielded
positive effects ($+0.275$ to $+0.560$; all
$p<0.001$). These results support T3 robustness
across backends and judges, but do not establish
overall simulator fidelity. Full details appear in
Appendix~\ref{sec:appendix-cross-crs}.
\section{Conclusion}
\label{sec:conclusion}

We presented \textit{SimInter}, a pipeline for LLM-based user simulation
that elicits qualitative behavioral attributes through structured
synthetic interviews and guards against fabrication via explicit
turn-level auditing. On $300$ ReDial users, simulators conditioned on
enriched profiles discuss movies along substantially richer behavioral
aspects than a statistics-only baseline ($d_z=0.96$,
$p_{\mathrm{adj}}<0.001$; zero-aspect users drop from $41\%$ to $5\%$);
pooled popularity alignment with MovieLens-25M reaches $r=0.347$; and
$94\%$ of enriched-attribute claims are human-supported. Enrichment
also concentrates item mentions and reduces lexical variety,
consistent with profile-conditioned generation narrowing expression
toward profile vocabulary. User-simulation quality in CRS is thus
multidimensional: the goal is not to maximize every diversity metric,
but to gain more on dimensions that matter for CRS evaluation than is
lost on those that do not.

\section*{Limitations}
\label{sec:limitations}

\paragraph{Evidence grounding and audit reliability.}
SimInter grounds enriched attributes in historical user
evidence, while independent validation against real
user behavior remains an important direction for
future work. Human evaluation assesses evidence
support rather than direct behavioral fidelity.
Moreover, audit decisions involve uncertainty,
particularly when distinguishing unsupported
rationalizations from legitimate preference
elaboration. Automatically retrieved evidence
and differences in annotator judgments further
limit conclusions about audit reliability.

\paragraph{Evaluation scope and generalizability.}
Our ablations do not include direct LLM-based
attribute inference without interviewing, leaving
the interview's independent contribution unresolved.
Sparse and imbalanced preference evidence limits
individual-level alignment evaluation. Enrichment
also involves trade-offs between aspect richness,
item diversity, and lexical variety. Although
cross-CRS experiments support the robustness of
aspect-level findings, they rely on automated
evaluation and do not establish agreement with
real-user assessments of recommenders.
Generalization beyond English movie recommendation
remains to be investigated.

\section*{Ethical Considerations}
ReDial is released under CC BY 4.0, MovieLens-25M is available for
non-commercial research use, and the iEvaLM-CRS implementations of
KBRD and UniCRS~\cite{wang-etal-2023-rethinking-evaluation} are under
the MIT License; we use all of them solely for research. LLMs are
accessed through their official APIs under the providers' terms.

\bibliography{custom}

\appendix

\section{T2 Degeneracy Diagnostics}
\label{sec:appendix-t2}

The main text reports only the pooled T2 popularity correlation 
(\S\ref{sec:eval-results}). This appendix documents why per-user and 
persona-anchored variants are not reportable.

\subsection{Per-User Popularity Correlation}

For each user, the per-user popularity Pearson $r$ is the correlation 
between the simulator's rating and the MovieLens-25M mean rating 
across the movies that user mentioned. This requires a minimum 
sample of $\sim 30$ (user, movie) pairs per user for a reliable 
estimate.

Only $6/300$ \textsc{Full} users reach this threshold (median $7$
pairs per user, maximum $47$); on the matched subset, only $2$ users in each
of \textsc{Full} and \textsc{StatisticalOnly}
reach this threshold. A paired test across users is therefore not
feasible. The pooled estimate in Table~\ref{tab:headline} aggregates
all (user, movie) pairs across the cohort, treating each pair as an
independent observation, a weaker statistical claim than per-user
testing would provide, but the only one available at our sample size.

\subsection{Persona-Anchored T2 (Liked vs.\ Disliked)}

A second variant of T2 correlates the simulator's in-conversation 
rating with the user's ReDial liked/disliked label (binary ground 
truth). Table~\ref{tab:appendix-t2-persona} reports the diagnostics.
In every condition, $94$--$97\%$ of high-confidence ground-truth labels
are \textit{liked}, giving $\text{variance}_{\text{redial}}<0.1$, a
near-constant variable. Pearson $r$ against a near-constant variable
is driven by a handful of disliked pairs and is not a reliable
discriminative metric, so the apparent ordering across conditions is
not interpreted.
\begin{table}[t]
\centering
\footnotesize
\setlength{\tabcolsep}{3.5pt}
\resizebox{\columnwidth}{!}{%
\begin{tabular}{lrr}
\toprule
 & \textbf{\textsc{Full}} & \textbf{\textsc{StatOnly}} \\
\midrule
Persona $r$ (pooled)        & 0.603  & 0.261  \\
$n$ pairs                   & 52     & 92     \\
Var.$_{\text{redial}}$      & 0.078  & 0.074  \\
\% liked (GT)               & 94.0\% & 96.3\% \\
$n$ high-conf.\ pairs       & 50     & 81     \\
$n$ users, both classes     & 2      & 2      \\
Accuracy (high-conf.)       & 94.0\% & 96.3\% \\
\bottomrule
\end{tabular}%
}
\caption{T2 persona-anchored diagnostics on the matched 100-user
subset. Ground-truth labels are near-constant in every condition,
making Pearson $r$ unreliable as a discriminative metric.}
\label{tab:appendix-t2-persona}
\end{table}
\subsection{Balanced Accuracy as an Alternative}

A natural alternative is balanced accuracy on liked/disliked 
classification. Computing this requires users with both classes in 
ground truth. As shown above, only $2$ users per condition
satisfy this, too few to estimate a meaningful
per-user balanced accuracy distribution. Pooled simple accuracy ($94$--$96\%$) equals
the majority-class rate and is uninformative.

\textbf{Recommendation for future work.} Per-user behavioral 
alignment metrics in CRS user simulation require either (a) a larger 
sample where most users have $\geq 30$ rated items in conversation, 
or (b) controlled prompting that elicits a balanced acceptance/ 
rejection ratio per user (e.g., $50/50$ targeted CRS recommendations). 
Both would address the per-user consistency gap noted by 
\citet{yoon2024evaluating} and \citet{bernard2025limitations}.

\section{Auditor Flag Taxonomy and Examples}
\label{sec:appendix-auditor}

\subsection{Flag Taxonomy}

\begin{table}[t]
\centering
\small
\begin{tabular}{lrrrr}
\toprule
\textbf{Issue type} & \textbf{High} & \textbf{Med.} & \textbf{Low} & \textbf{Tot.} \\
\midrule
fabrication            & 12 & 47  & 72  & 131 \\
caricature             & 12 & 255 & 105 & 372 \\
history\_contradiction & 55 & 68  & 50  & 173 \\
structural\_error      & 0  & 0   & 0   & 0   \\
\midrule
\textbf{Total}         & 79 & 370 & 227 & 676 \\
\bottomrule
\end{tabular}
\caption{Distribution of auditor flags by type and severity across 
$300$ interviews (final attempts). Dialogue-level decision
distribution: \textsc{accept} $209$, \textsc{partial} $91$,
\textsc{regenerate} $0$ (one interview regenerated once before
receiving \textsc{partial}).}
\label{tab:appendix-flags}
\end{table}

Table~\ref{tab:appendix-flags} summarizes the full flag distribution 
across the $300$ interviews. We make three observations:

\textbf{1. Invented rationalization is the dominant pattern} ($372/676$
flags are \textit{caricature}, $69\%$ of them medium severity): the
simulator frequently produces \emph{reasons} for preferences that
sound plausible but are not attested in the profile.

\textbf{2. Flags grounded in the user's own evidence are the most
reliable.} Contradictions with the liked and disliked lists or with
the ReDial conversation can be checked directly (Examples 2--4).
Fabrication flags about film facts rely on the auditor's parametric
knowledge and are less reliable: in several cases the auditor flagged
accurate scene descriptions as fabricated. Such errors are conservative: they can remove grounded content from
the enrichment input, but they do not introduce unsupported claims
into the profile.

\textbf{3. \textsc{regenerate} is rare.} The auditor reserves
\textsc{regenerate} for structural failure or $\geq 3$ high-severity
flags. This occurred in one of $300$ interviews, and the regenerated
interview received \textsc{partial}, indicating that gross structural
failure is rare with capable LLMs (Gemini 2.5 Flash) at our interview
scale. The dominant failure mode is local rather than global.

\subsection{Representative Example 1: \textsc{accept}}
\label{sec:appendix-example-accept}

User 78. The auditor raised no flags; every film the interviewee
mentions appears in the user's liked or disliked lists.

\begin{quote}\small
\textbf{Q1:} If a friend asked you to recommend a movie that captures
your taste, what would you pick and why?

\textbf{A1:} I'd probably point them toward something like \textit{Juno}
or \textit{The Perks of Being a Wallflower}. I really lean toward
comedies and dramas that have a sharp, witty sense of humor while
still hitting you with some genuine, deeper moments.

\textbf{Q4:} [...] When does that attempt at honesty in a drama or
comedy feel unearned or fall flat for you?

\textbf{A4:} [...] Like when I watched \textit{White Noise}, it just
didn't land for me at all---it missed that grounded, witty balance and
felt a bit too forced.
\end{quote}

\subsection{Representative Example 2: \textsc{partial} (history contradiction)}
\label{sec:appendix-example-history}

User 1. One high-severity \textit{history\_contradiction} flag (turn~6)
and one low-severity \textit{caricature} flag (turn~10); turn~6 was
excluded from enrichment.

\begin{quote}\small
\textbf{A (turn 6, excluded):} I just really enjoy that buddy-cop or
mismatched-duo comedy style [...]. It's the same kind of fun,
fast-paced energy I liked in \textit{Blue Streak} with Martin
Lawrence [...]
\end{quote}

\textbf{Auditor explanation:} ``The user claims to have liked
\textit{Blue Streak}'s energy in the recent dialogue, but in the
earlier conversation explicitly stated they had not seen
\textit{Blue Streak} and expressed intent to watch it in the future.''

\subsection{Representative Example 3: \textsc{partial} (caricature)}
\label{sec:appendix-example-caricature}

User 730. One high-severity \textit{caricature} flag (turn~8), which
was excluded from enrichment.

\begin{quote}\small
\textbf{A (turn 8, excluded):} Honestly, a lot of big-budget stuff just
gets way too chaotic and messy without really earning those stakes,
like some of the massive CGI battles in \textit{Justice League} where
it's hard to actually care about what's happening.
\end{quote}

\textbf{Auditor explanation:} ``The user claims \textit{Justice
League}'s CGI battles don't earn stakes and are hard to care about,
but \textit{Justice League} is in their liked movies list with no
negative commentary in the profile.'' This illustrates an
\emph{invented rationalization}: a plausible reason for a dislike that
the user's history does not support.
\subsection{Representative Example 4: \textsc{partial} (invented rationalization)}
\label{sec:appendix-example-caricature-multi}

User 66. One high-severity \textit{caricature} flag (turn~8) and one
medium-severity \textit{caricature} flag (turn~2); both turns were
excluded from enrichment.

\begin{quote}\small
\textbf{A (turn 8, excluded):} I think it fails when a comedy tries way
too hard to be funny without caring about the characters, like in
\textit{The Hangover} or \textit{Superbad}. It just feels loud and
obnoxious rather than having any real situation driving the humor.
\end{quote}

\textbf{Auditor explanation:} ``This response performs a snobby
cinephile persona dismissing comedies as `loud and obnoxious,' but the
profile shows the user actually likes comedies like \textit{Super
Troopers}, \textit{Beerfest}, and \textit{Tommy Boy}.''

Both films are on the user's disliked list, so the dislike itself is
attested; the stated reason is not. The simulator invents a
plausible justification that is inconsistent with the user's liking
for comparably broad comedies---an \emph{invented rationalization} in
the sense of Stage~4.

\section{Per-User Metric Distributions}
\label{sec:appendix-per-user}

The main text (\S\ref{sec:eval-results}) shows the per-user T3 aspect 
entropy distribution. This appendix shows the corresponding 
distributions for all other primary metrics on the matched 100-user
subset. Throughout, \textsc{Full} is dark blue and
\textsc{StatisticalOnly} is light blue, with dashed lines marking
condition means.

\begin{figure}[t]
\centering
\includegraphics[width=\columnwidth]{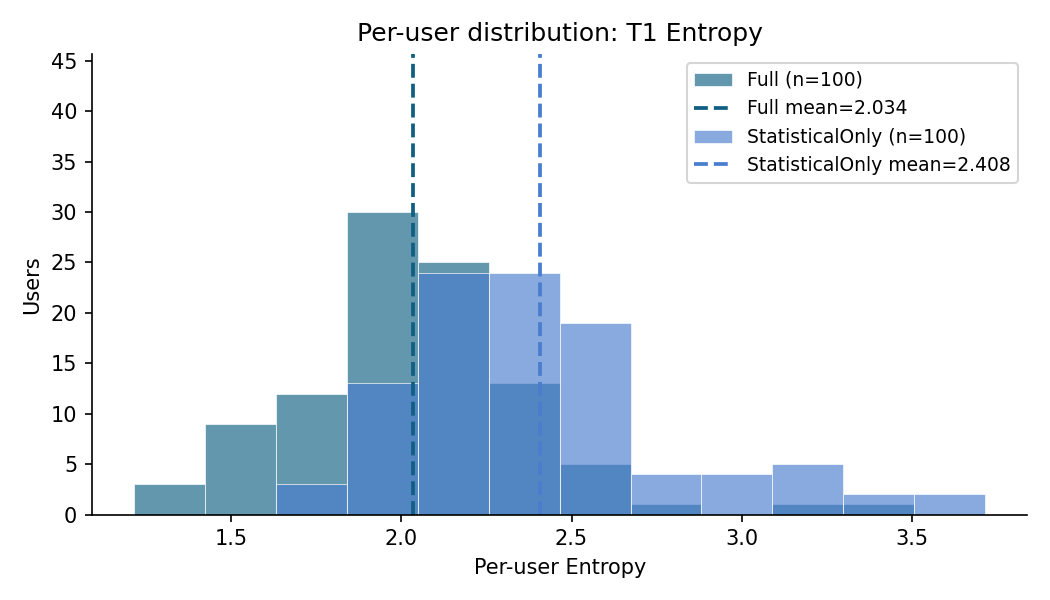}
\caption{T1 per-user item entropy. \textsc{Full} is shifted left of
\textsc{StatisticalOnly} ($2.034$ vs.\ $2.409$; $\Delta=-0.375$,
$d_z=-0.95$, $p_{\mathrm{adj}}<0.001$), reflecting a more concentrated
movie-mention distribution.}
\label{fig:appendix-t1-entropy}
\end{figure}

\begin{figure}[t]
\centering
\includegraphics[width=\columnwidth]{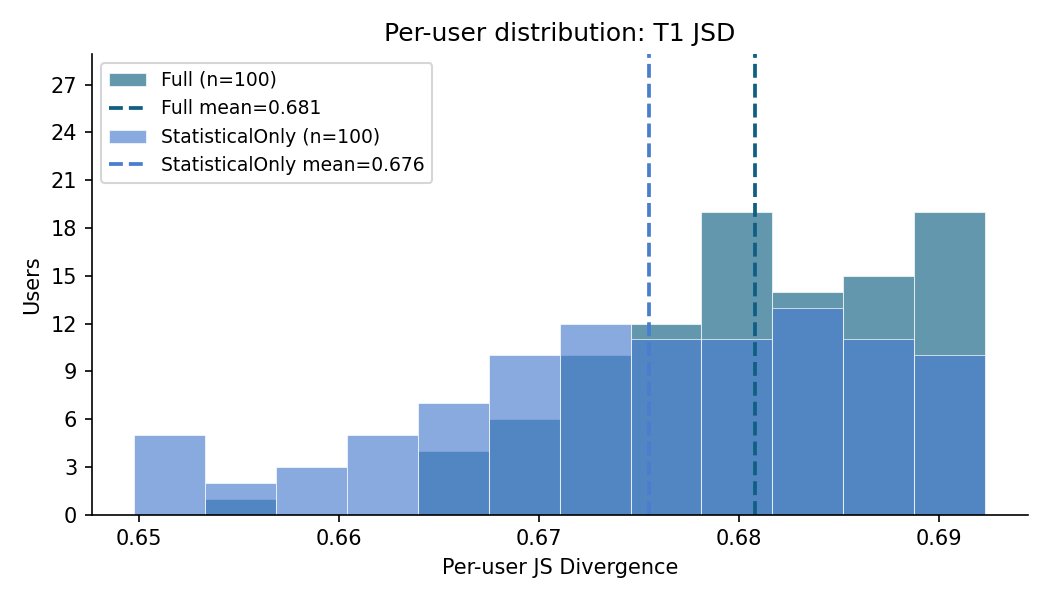}
\caption{T1 per-user JSD against the ReDial reference distribution.
\textsc{Full} is marginally higher ($0.681$ vs.\ $0.676$; $d_z=0.47$,
$p_{\mathrm{adj}}<0.001$); the absolute difference is small.}
\label{fig:appendix-t1-jsd}
\end{figure}

\begin{figure}[t]
\centering
\includegraphics[width=\columnwidth]{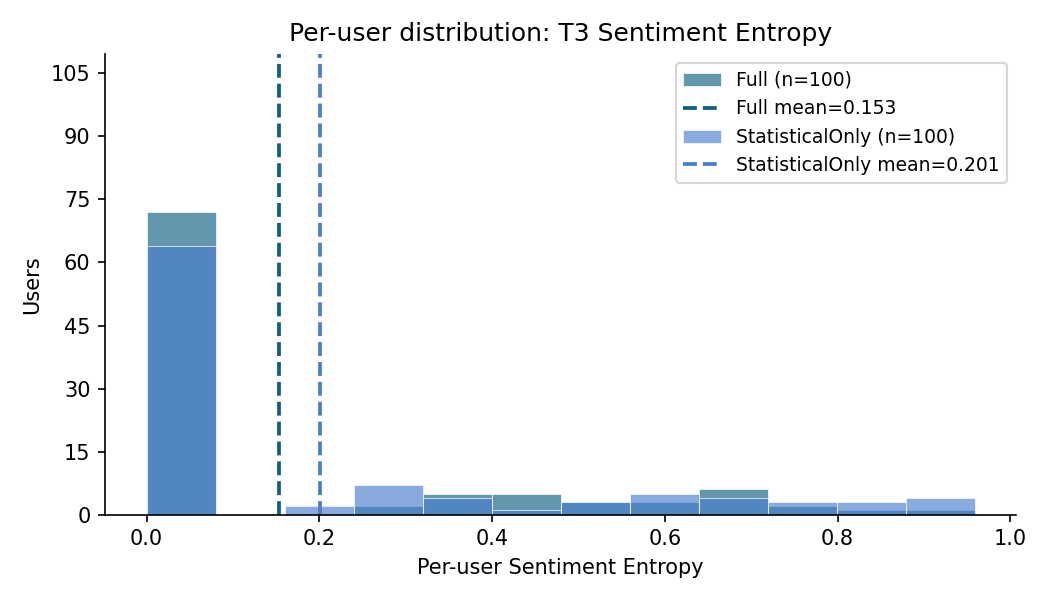}
\caption{T3 per-user sentiment entropy. \textsc{Full} is slightly
lower ($0.153$ vs.\ $0.201$; $d_z=-0.26$, uncorrected $p=0.011$). Both
conditions cluster at zero ($72$ and $64$ users), reflecting a strong
positive-sentiment imbalance.}
\label{fig:appendix-t3-sentiment}
\end{figure}

\begin{figure}[t]
\centering
\includegraphics[width=\columnwidth]{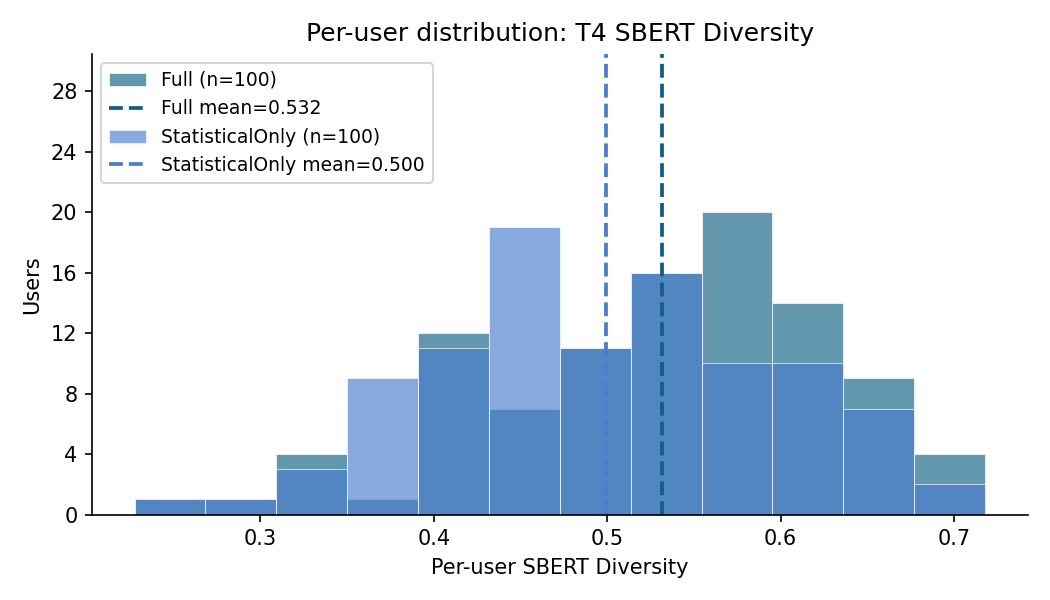}
\caption{T4 per-user SBERT semantic diversity. \textsc{Full} is
slightly higher ($0.532$ vs.\ $0.500$; $d_z=0.25$,
$p_{\mathrm{adj}}=0.11$); $10$ \textsc{Full} and $15$
\textsc{StatisticalOnly} users fall below $0.4$.}
\label{fig:appendix-t4-sbert}
\end{figure}

\begin{figure}[t]
\centering
\includegraphics[width=\columnwidth]{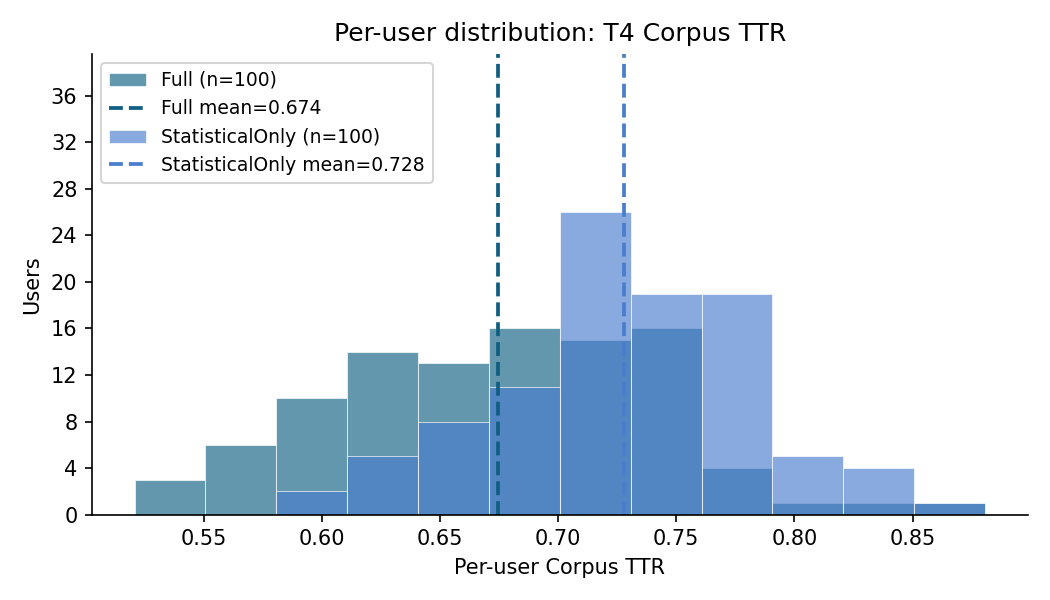}
\caption{T4 per-user corpus TTR. \textsc{Full} TTR is systematically
lower ($0.674$ vs.\ $0.728$; $d_z=-0.68$, uncorrected $p<0.001$);
enriched users repeat tokens more.}
\label{fig:appendix-t4-ttr}
\end{figure}

\begin{figure}[t]
\centering
\includegraphics[width=\columnwidth]{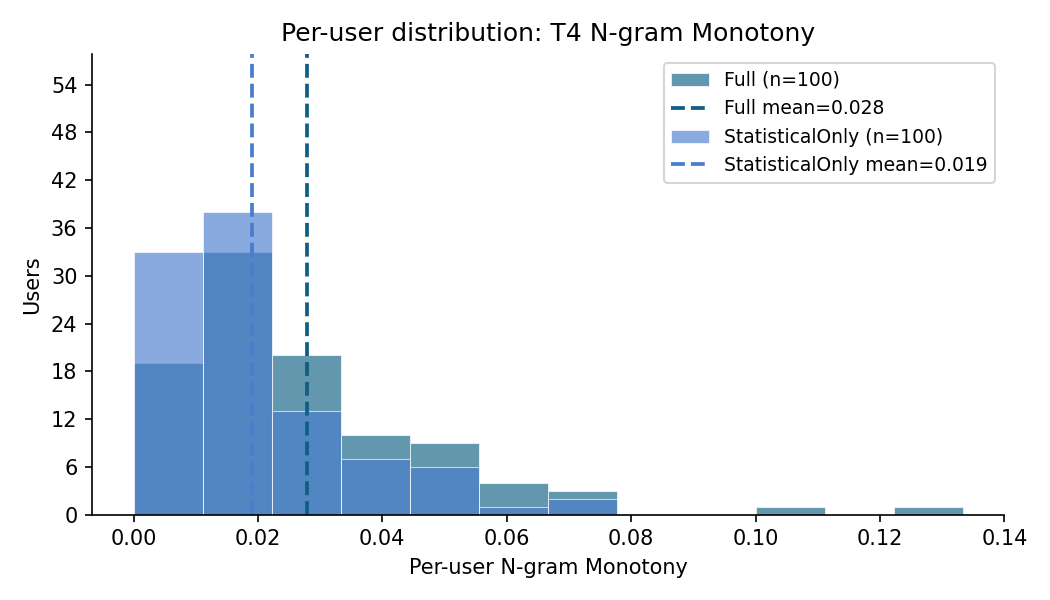}
\caption{T4 per-user n-gram monotony (Jaccard overlap of 
bigrams/trigrams between turn pairs; higher = more repetitive). 
\textsc{Full} is skewed right of \textsc{StatisticalOnly} ($0.028$
vs.\ $0.019$; $d_z=0.36$, uncorrected $p<0.001$), reflecting
profile-vocabulary reuse.}
\label{fig:appendix-t4-monotony}
\end{figure}

\section{T3 Aspect Vocabulary}
\label{sec:appendix-t3}

\begin{table*}[t]
\centering
\small
\begin{tabular}{rlrlr}
\toprule
\textbf{Rank} & \textbf{\textsc{Full} aspect} & \textbf{Count} & \textbf{\textsc{StatOnly} aspect} & \textbf{Count} \\
\midrule
1  & mood               & 164 & movie overall   & 407 \\
2  & pacing             & 103 & plot            & 58 \\
3  & movie overall      & 81  & mood            & 57 \\
4  & atmosphere         & 78  & premise         & 23 \\
5  & overall vibe       & 50  & action          & 22 \\
6  & plot               & 43  & overall concept & 21 \\
7  & vibe               & 37  & humor           & 20 \\
8  & overall experience & 27  & genre           & 20 \\
9  & humor              & 26  & comedy          & 18 \\
10 & mood/atmosphere    & 20  & intensity       & 12 \\
11 & energy             & 18  & overall fit     & 10 \\
12 & action             & 17  & drama           & 9 \\
13 & overall quality    & 14  & genre/mood      & 8 \\
14 & overall concept    & 14  & vibe            & 8 \\
15 & tone               & 11  & tone            & 7 \\
16 & overall appeal     & 11  & atmosphere      & 7 \\
17 & premise            & 11  & soundtrack      & 7 \\
18 & mood/vibe          & 11  & pacing          & 6 \\
19 & intensity          & 11  & thriller        & 6 \\
20 & style              & 10  & overall quality & 6 \\
\midrule
\multicolumn{2}{l}{Distinct aspects (filtered)} & \textbf{267} & & \textbf{174} \\
\bottomrule
\end{tabular}
\caption{Top-20 most frequent aspects on the matched 100-user subset
(after filtering generic placeholders). \textsc{Full}'s vocabulary
skews toward experiential dimensions (\textit{mood}, \textit{pacing},
\textit{atmosphere}, \textit{energy});
\textsc{StatisticalOnly} is dominated by the near-generic
\textit{movie overall} ($43\%$ of its aspect mentions vs.\ $7\%$ for
\textsc{Full}) and broad genre labels. Total distinct non-generic
aspects: $267$ vs.\ $174$.}
\label{tab:appendix-t3-aspects}
\end{table*}

Table~\ref{tab:appendix-t3-aspects} shows the top-20 aspects in each
condition after filtering generic placeholders. Beyond the aggregate
counts, the qualitative difference is informative: \textsc{Full}
introduces experiential vocabulary (\textit{pacing}, \textit{overall
vibe}, \textit{energy}) that is rare or absent in
\textsc{StatisticalOnly}, whose mentions concentrate on a single
holistic judgment (\textit{movie overall}).

\section{Cross-CRS Robustness with Trained Recommenders}
\label{sec:appendix-cross-crs}

We conducted a separate paired robustness experiment using the frozen
ABLATION-120 cohort, sampled from the original FULL-300 population
before inspecting the results. Each user interacted with KBRD and
UniCRS under two conditions: \textsc{Full}, containing the statistical
profile and semantic enrichment, and \textsc{StatisticalOnly},
containing the same statistical profile and pickiness score but no
semantic enrichment.

For every user, condition, and recommender, we generated four
conversations corresponding to the aspect, movie, mood, and expertise
opening strategies. This produced
$120 \times 2 \times 2 \times 4 = 1{,}920$ conversations. User
identity, behavioral statistics, opening-strategy schedule,
conversation budget, and recommender configuration were held fixed
across paired conditions.

\paragraph{Recommender implementations and generation.}
We used the iEvaLM-CRS implementations and released ReDial
checkpoints. KBRD combines a relational-graph recommender with a
BART-base conversation model (checkpoint files
\texttt{kbrd\_rec\_redial/model.pt} and
\texttt{kbrd\_conv\_redial/pytorch\_model.bin}; SHA-256 prefixes
\texttt{d62f7548} and \texttt{dd85488a}). UniCRS combines
knowledge-graph prompts with DialoGPT-small and a RoBERTa-base text
encoder (\texttt{unicrs\_rec\_redial/model.pt} and
\texttt{unicrs\_conv\_redial/model.pt}; SHA-256 prefixes
\texttt{9cd55118} and \texttt{23599951}). For UniCRS, the released
checkpoint's padding-token index was mapped to the \texttt{<movie>}
placeholder before movie-title substitution. Both recommenders used a
maximum response length of 128 tokens; UniCRS additionally used a
no-repeat 3-gram constraint. The user simulator was Gemini 2.5 Flash
with temperature $0.7$, a 256-token output limit, and reasoning
disabled. Each conversation was generated once without a fixed
sampling seed.

Unlike the primary experiment, which used Claude Haiku 4.5 for
evaluation, this robustness experiment evaluated the same stored
conversations with separate GPT-4o-mini and DeepSeek-chat calls, both
at temperature $0.0$. Each evaluator used its own extraction outputs
and T3 reference baseline (baseline fingerprints
\texttt{1afc994dd78c1045} and \texttt{82a23731b46cf48d}, respectively).
We therefore interpret within-judge paired differences and do not
compare absolute entropy values across judges or with the primary
experiment.

\paragraph{Post-hoc CRS-dialogue audit.}
This Phase-2 audit is distinct from the Stage~4 Audit Agent that
audits interviews before enrichment. GPT-4o-mini (temperature $0.2$)
checked the simulated user's messages for fabrication, caricature,
role confusion, and persona drift, with a 512-token output limit. It
assigned \textsc{Accept} when there was no major issue (at most a minor
issue in one turn), \textsc{Partial} when one issue type was present
but the conversation remained mostly usable, and \textsc{Reject} when
multiple issue types or one severe issue were present; role confusion
across multiple turns was explicitly severe. Model, prompt,
temperature, and criteria were fixed across recommenders and
conditions; only the condition-specific profile and generated
conversation differed. Audit decisions were produced before T3
extraction and without access to extraction outputs. The audit affected
inclusion only and did not regenerate conversations. Although the
auditor and one T3 judge both used GPT-4o-mini, they were separate calls
with different prompts; the DeepSeek-chat evaluation used a different
underlying model.

\paragraph{Audit filtering and paired-sample construction.}
All four CRS--condition arms contained 480 technically completed
conversations (120 users $\times$ four opening strategies), and no
conversation terminated with a transport or generation error. In the
primary analysis, individual conversations assigned a Phase-2
\textsc{Reject} decision were excluded, whereas \textsc{Accept} and
\textsc{Partial} conversations were retained. Rejecting one
conversation did not remove the entire user: a user remained evaluable
in an arm whenever at least one of their four conversations was
retained. The paired analysis then used the intersection of users
evaluable in both conditions. This produced 111 paired users for KBRD
(112 evaluable in \textsc{Full} and 118 in
\textsc{StatisticalOnly}) and 80 for UniCRS (94 and 99,
respectively).

The conversation-level rejection rates were 39.6\% for
\textsc{Full} and 36.5\% for \textsc{StatisticalOnly} with KBRD,
and 68.5\% and 61.7\%, respectively, with UniCRS. A post-hoc paired
Wilcoxon diagnostic over per-user rejection rates found no significant
condition difference for KBRD ($p=0.262$), but found a higher rejection
rate for \textsc{Full} with UniCRS ($p=0.019$; two-sided). We therefore report
the audit-filtered analysis together with an all-conversation
sensitivity analysis.

\paragraph{Statistical inference.}
For each recommender--judge combination, inference was performed on
paired user-level differences. We report percentile 95\% confidence
intervals from 10,000 paired bootstrap resamples (seed $0$), two-sided
Wilcoxon signed-rank tests with zero differences removed and SciPy's
automatic exact/asymptotic method selection, and paired Cohen's
$d_z$. The table reports unadjusted $p$-values for these four T3
comparisons. As a robustness check, two-sided paired sign-flip tests with
100,000 randomizations (seed $1$) gave the same significance
conclusions.

\begin{table*}[!t]
\centering
\scriptsize
\setlength{\tabcolsep}{4pt}
\begin{tabular}{llrrrrrrr}
\toprule
\textbf{CRS} & \textbf{Evaluation judge} & \textbf{$n_u$}
& \textbf{\textsc{Full}} & \textbf{\textsc{StatOnly}}
& \textbf{$\Delta$ T3} & \textbf{95\% CI}
& \textbf{$p_{\mathrm{W}}$} & \textbf{$d_z$} \\
\midrule
KBRD   & GPT-4o-mini   & 111 & 0.392 & 0.126
       & +0.265 & [0.144, 0.389] & 0.0002 & 0.40 \\
KBRD   & DeepSeek-chat & 111 & 0.350 & 0.069
       & +0.280 & [0.170, 0.394] & $<0.0001$ & 0.46 \\
UniCRS & GPT-4o-mini   & 80 & 0.488 & 0.157
       & +0.330 & [0.179, 0.487] & 0.0001 & 0.47 \\
UniCRS & DeepSeek-chat & 80 & 0.498 & 0.129
       & +0.369 & [0.214, 0.527] & $<0.0001$ & 0.52 \\
\bottomrule
\end{tabular}
\caption{Primary audit-filtered Cross-CRS results for per-user filtered
T3 aspect entropy. Here, $n_u$ is the number of users with at least one
\textsc{Accept} or \textsc{Partial} conversation in both conditions.
For each user and condition, aspect--sentiment pairs were pooled across
all retained conversations before computing entropy.
Differences are
\textsc{Full}$-$\textsc{StatisticalOnly};
$p_{\mathrm{W}}$ denotes the two-sided paired Wilcoxon signed-rank
test. All bootstrap confidence intervals exclude zero. Because means
and differences are computed from unrounded user-level values, a
displayed difference need not equal the subtraction of the two rounded
means. Absolute entropy values should not be compared across evaluation
judges.}

\label{tab:cross-crs-t3}
\end{table*}

\begin{figure}[!t]
  \centering
  \includegraphics[width=\columnwidth]{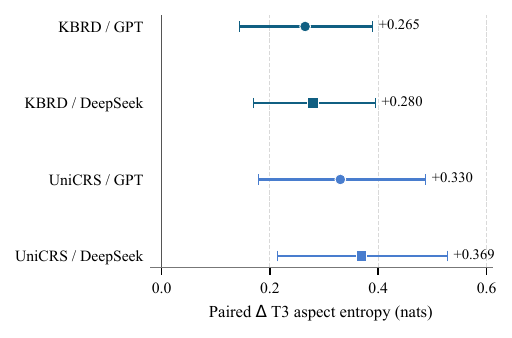}
  \caption{Paired effect of semantic enrichment on filtered T3 aspect
  entropy. Points represent mean user-level differences and horizontal
  bars show 95\% bootstrap confidence intervals. The vertical line
  marks the null effect. Circles denote GPT-4o-mini and squares denote
  DeepSeek-chat; dark and light blue denote KBRD and UniCRS.}
  \label{fig:cross-crs-t3}
\end{figure}

\paragraph{Per-user T3 aggregation.}
For each user and condition, the evaluator extracted aspect--sentiment
pairs from every retained conversation and merged them into a single
user-level aspect-frequency distribution. Filtered T3 aspect entropy
was then computed once from this merged distribution. We did not
compute a separate entropy for each opening strategy and then average
the four values. The paired effect was calculated as
$\Delta_u=H_u^{\textsc{Full}}-H_u^{\textsc{StatisticalOnly}}$, and
statistical inference was performed over these user-level differences.

Because retained conversation counts could differ between conditions,
the filtered entropy estimates may be sensitive to unequal exposure.
Among paired users, the mean numbers of retained conversations were
2.60 for \textsc{Full} and 2.62 for \textsc{StatisticalOnly} with
KBRD, and 1.66 and 1.80, respectively, with UniCRS. Pooling more
conversations can expose more aspect categories and thereby affect
entropy, so the filtered estimates may reflect both condition effects
and retained-exposure differences.
We therefore use the all-conversation analysis below, which retains all
four opening strategies in both conditions for all 120 users, as a
sensitivity check.

Table~\ref{tab:cross-crs-t3} and
Figure~\ref{fig:cross-crs-t3} show a positive primary T3 effect for
both recommenders and both evaluation judges. The all-conversation
sensitivity analysis yielded effects of
$+0.275$ and $+0.371$ for KBRD, and $+0.560$ and $+0.534$ for
UniCRS, under GPT-4o-mini and DeepSeek-chat, respectively
(all $p<0.001$). Here, ``technically completed'' means that a complete
conversation record was produced without a transport or generation
error; it does not require natural acceptance and includes
max-turn terminations.

Per-user T3 scores were strongly correlated between the two evaluation judges: Spearman's $\rho=0.77$ in the primary analysis and
$\rho=0.72$ in the all-conversation sensitivity analysis. The
consistent direction makes unequal exposure an unlikely sole
explanation of the T3 result, but does not establish overall simulator
quality or fidelity to individual real users.

\section{Human Evaluation Protocol}
\label{sec:appendix-human-eval}

\subsection{Annotation Task}

Each annotation task presented an annotator with a single 
\textbf{claim} (one of the six attribute values from a 
\textsc{Full}-enriched user profile), the user's 
\textbf{behavioral profile} (Stage 1--2 statistics), and the user's 
\textbf{interview transcript} (Stage 3 output, including any turns 
the auditor had flagged). Annotators were asked to judge whether the 
claim was supported by the interview evidence.

\subsection{Rating Scales}

\textbf{Support} (4-class):
\begin{itemize}
\item \textit{Supported}: the claim is directly justified by specific 
content in the interview.
\item \textit{Partially supported}: the interview contains relevant 
content but does not fully justify the specific claim (e.g., the 
claim is more specific or stronger than the evidence).
\item \textit{Unsupported}: the interview does not contain evidence 
relevant to the claim, even though the claim is consistent with the 
profile.
\item \textit{Contradicted}: the interview contains evidence 
inconsistent with the claim.
\end{itemize}

\textbf{Grounding} (3-class):
\begin{itemize}
\item \textit{Evidence-grounded}: the claim is anchored to a specific 
quotation or behavioral pattern in the interview.
\item \textit{Generic}: the claim is consistent with the interview 
but could apply to many users; the evidence is weak or non-specific.
\item \textit{Cliche/stereotype}: the claim performs a recognizable 
persona archetype not specifically attested in the user's data.
\end{itemize}

\textbf{Numeric ratings} (1--5):
\begin{itemize}
\item \textit{Consistency}: how consistent the claim is with the 
user's behavioral profile (1 = inconsistent, 5 = highly consistent).
\item \textit{Usefulness}: how useful the claim would be for 
conditioning a simulator (1 = not useful, 5 = highly useful).
\item \textit{Overreach}: how much the claim overshoots the available 
evidence (1 = no overreach, 5 = severe overreach).
\item \textit{Confidence}: the annotator's confidence in their 
judgment (1 = uncertain, 5 = highly confident).
\end{itemize}

Mean ratings across all $315$ annotations are $3.77$ (consistency),
$3.61$ (usefulness), $1.70$ (overreach), and $3.00$ (confidence).

\subsection{Annotator Pool and Procedure}

Three annotators (graduate students with prior NLP annotation 
experience) independently rated claims sampled at random from the
$582$ enriched-attribute claims of the \textsc{Full} profiles. We
report the $105$ claims, drawn from $45$ users and covering all six
attribute types, that were rated by all three annotators ($315$
annotations). Annotators worked independently without discussion of
disagreements; no calibration round was performed prior to
annotation, a methodological choice we identify as a limitation
(see Limitations).

\subsection{Per-Attribute Detail}

\begin{table}[t]
\centering
\small
\setlength{\tabcolsep}{4pt}
\begin{tabular}{lrrrr}
\toprule
\textbf{Attribute} & \textbf{Sup.} & \textbf{Part.} & \textbf{Uns.} & \textbf{Contr.} \\
\midrule
Viewing motivation        & 27.5\% & 72.5\% & 0.0\%  & 0.0\%  \\
Aspect prefs.\ (liked)    & 75.0\% & 25.0\% & 0.0\%  & 0.0\%  \\
Aspect prefs.\ (descr.)   & 33.3\% & 66.7\% & 0.0\%  & 0.0\%  \\
Pickiness                 & 72.7\% & 24.2\% & 0.0\%  & 3.0\%  \\
Domain expertise          & 27.8\% & 61.1\% & 11.1\% & 0.0\%  \\
Novelty preference        & 26.7\% & 60.0\% & 13.3\% & 0.0\%  \\
Aspect prefs.\ (disliked) & 38.1\% & 42.9\% & 9.5\%  & 9.5\%  \\
Decision-making style     & 11.9\% & 66.7\% & 21.4\% & 0.0\%  \\
\midrule
\textbf{All}              & 45.1\% & 48.3\% & 5.4\%  & 1.3\%  \\
\bottomrule
\end{tabular}
\caption{Full per-attribute breakdown of support ratings
(annotation level). The $1.3\%$ \textit{Contradicted} rate
concentrates in disliked aspects ($9.5\%$) and \textit{pickiness}
($3.0\%$); \textit{decision-making style} has the highest
\textit{Unsupported} rate ($21.4\%$), consistent with cognitive-style
claims being the most amenable to invented rationalization.}
\label{tab:appendix-human-eval-detail}
\end{table}

\subsection{Inter-Annotator Agreement Details}

We report multiple agreement metrics. Since the Support label is 
ordinal (Supported $\succ$ Partially supported $\succ$ Unsupported 
$\succ$ Contradicted), quadratic-weighted $\kappa$ is the 
methodologically appropriate primary metric~\citep{cohen1968weighted}: 
it penalizes large ordinal disagreements (Supported vs.\ Contradicted) 
more than small ones (Supported vs.\ Partially supported).

\paragraph{Support label (4-class, ordinal).} Pairwise agreement 
across the $105$ triply-annotated claims:

\begin{table}[h]
\centering
\small
\begin{tabular}{lrrrr}
\toprule
\textbf{Pair} & \textbf{Unwgt. $\kappa$} & \textbf{Wgt. $\kappa$} & \textbf{Exact} & \textbf{$\pm 1$-rank} \\
\midrule
A vs.\ B & 0.30 & 0.43 & 60.0\% & 98.1\% \\
A vs.\ C & 0.19 & 0.40 & 45.7\% & 97.1\% \\
B vs.\ C & 0.43 & 0.52 & 69.5\% & 98.1\% \\
\bottomrule
\end{tabular}
\caption{Pairwise inter-annotator agreement on the Support label. 
Weighted $\kappa$ is quadratic-weighted; ``$\pm 1$-rank'' is the 
rate at which two annotators fell in the same or adjacent ordinal 
category. Annotator C applies the strictest interpretation of
Supported.}
\label{tab:appendix-iaa-support}
\end{table}

\paragraph{Grounding label (3-class, nominal).} For the Grounding 
label (Evidence-grounded / Generic / Cliche-stereotype), $\kappa$ 
values are: A vs.\ B $= 0.12$, A vs.\ C $= 0.27$, B vs.\ C 
$= 0.37$. Agreement on grounding is fair, mainly reflecting the
boundary between Generic and Evidence-grounded; only one of $315$
annotations used Cliche/stereotype.

\paragraph{Binary support.} Collapsing the 4-class label into 
positive (Supported $+$ Partially supported) vs.\ negative 
(Unsupported $+$ Contradicted) gives a coarser agreement signal:

\begin{itemize}
\item A vs.\ B: $\kappa = 0.20$, raw agreement $93.3\%$
\item A vs.\ C: $\kappa = 0.37$, raw agreement $89.5\%$
\item B vs.\ C: $\kappa = 0.11$, raw agreement $88.6\%$
\end{itemize}

All three annotators agree on $85.7\%$ of claims under this split.
Binary $\kappa$ is depressed by the high prevalence of positive labels
(the kappa paradox); Gwet's AC1~\citep{Gwet2008ComputingIR}, which is
robust to prevalence, is $0.89$ for this split, and $0.99$ for
Contradicted versus all other labels.

\paragraph{Disagreement pattern.} The most consistent disagreement 
is between Supported and Partially supported: in $32\%$ of annotator
pairs per claim, one annotator marked the claim Supported while the
other marked it Partially supported. This boundary was under-specified
in our protocol: the task statement did not operationalize how much
explicit textual evidence was required to justify the stronger
rating.

\paragraph{Annotator C as strictest rater.} Annotator C marks claims
Supported far less often than A and B ($24.8\%$ vs.\ $68.6\%$ and
$41.9\%$), while all three judge most claims at least partially
supported ($88.6\%$, $93.3\%$, and $98.1\%$ for C, A, and B). Manual
inspection suggests C required near-verbatim evidence for Supported,
whereas A and B accepted paraphrased evidence. We retain all three
annotators in the main results.

\paragraph{What the agreement numbers mean for the paper's claims.} 
The $94\%$ aggregate support rate quoted in \S\ref{sec:human-eval} 
is robust under several reasonable agreement assumptions: (a) it 
remains $\geq 91\%$ under any single-annotator-out leave-one-out; 
(b) within-one-rank agreement of $97$--$98\%$ across all annotator 
pairs indicates strong convergence on the direction of support, 
which is the quantity the headline claim relies on; (c) the 
$1.3\%$ Contradicted rate, where chance-corrected agreement is
highest (AC1 $=0.99$), is the single most claim-critical number for
paper integrity (it bounds the fabrication rate the auditor failed to
catch). A revised protocol with sharper category definitions, an
agreement-calibration pre-annotation phase, and a larger pool of
annotators is the natural next step for human-eval methodology in
this domain.

\section{Human Evaluation Protocol for Audit Decisions}
\label{sec:appendix-human-audit}

\paragraph{Evaluation material and sampling.}
The evaluation unit was a single synthetic interviewee response
produced by the SimInter user-simulation pipeline and processed by
Stage~4 in the $300$-user experiment. These are \emph{generated}
interview responses, not utterances taken from ReDial. Each response
was linked to the historical ReDial user whose interactions informed
the simulation. We selected $105$ turns from distinct users, equally
stratified by the auditor's final outcome: $35$ \textsc{Excluded},
$35$ \textsc{Flagged--Retained}, and $35$ \textsc{Unflagged}.
The three groups were presented as consecutive $35$-case blocks in a
fixed order shared by all annotators, but the group boundaries and
auditor outputs were concealed.

\paragraph{Annotation procedure and human reference.}
Three co-author annotators independently reviewed the same $105$ turns
($315$ judgments in total). They saw the target synthetic response,
optional preceding synthetic-interview context, automatically retrieved
statements from the matching historical ReDial user's original
conversations, and optional derived statistical-profile information.
They did not see the auditor's action, issue flags, or explanations.
The three judgment categories were interpreted as follows:
\textsc{Excluded}, an evidence-grounded problem warrants removing the
entire synthetic turn; \textsc{Flagged--Retained}, a meaningful concern
exists but does not warrant removal; and \textsc{Unflagged}, no meaningful
issue is identified. We used the majority vote of the three
annotators as the human reference for each turn. The evaluation unit
and reference are at the \emph{whole-turn} level; they do not validate
the correctness of individual auditor flag types.

\paragraph{Inter-annotator agreement.}
All $105$ cases received three labels. Agreement was unanimous on
$70/105$ ($66.7\%$) and $2$--$1$ on the remaining $35$; there were
no three-way splits. Nominal Fleiss' $\kappa$ was $0.664$ (95\%
stratified case-bootstrap interval $[0.572,0.751]$; $2{,}500$
replicates, resampling within the three auditor strata). Within-stratum
Fleiss' $\kappa$ values were $0.337$ (excluded), $0.308$
(flagged-but-retained), and $0.268$ (unflagged). These lower
within-stratum values suggest that aggregate agreement is influenced
by the differences in label distributions across strata.

\begin{table}[t]
\centering
\small
\setlength{\tabcolsep}{4pt}
\begin{tabular}{@{}lrrrr@{}}
\toprule
\textbf{Auditor outcome} & \textbf{$n$} & \textbf{Human}
& \textbf{Human} & \textbf{Human} \\
 & & \textbf{Excluded} & \textbf{Flagged--ret.}
 & \textbf{Unflagged} \\
\midrule
Excluded          & 35 & 31 & 4  & 0  \\
Flagged--retained & 35 & 0  & 31 & 4  \\
Unflagged         & 35 & 1  & 5  & 29 \\
\midrule
\textbf{Total} & \textbf{105} & \textbf{32}
& \textbf{40} & \textbf{33} \\
\bottomrule
\end{tabular}
\caption{Auditor outcomes versus the majority judgments of three
annotators for $105$ synthetic interviewee turns. The diagonal
corresponds to $91/105$ ($86.7\%$) three-class human-referenced
accuracy on the balanced sample.}
\label{tab:human-audit}
\end{table}

\paragraph{Three-class and filtering accuracy.}
Three-class accuracy against the human-majority reference was
$91/105$ ($86.7\%$): $31/35$ ($88.6\%$) in the excluded stratum,
$31/35$ ($88.6\%$) in the flagged-but-retained stratum, and
$29/35$ ($82.9\%$) in the unflagged stratum
(Table~\ref{tab:human-audit}). Fourteen cases disagreed at the
three-class level: four auditor-excluded turns were judged
flagged-but-retained; four flagged-but-retained turns were judged
unflagged; and among auditor-unflagged turns, five were judged
flagged-but-retained and one excluded. Thus, nine of the fourteen disagreements concerned the distinction between flagged-but-retained and unflagged outcomes, without changing the final filtering action.

When both retained categories were collapsed to a single
\textsc{Retain} action, the auditor agreed with the majority human
filtering action on $100/105$ turns ($95.2\%$): $31/35$ auditor-excluded
turns and $69/70$ auditor-retained turns. On the same evaluation set,
an always-retain baseline would agree with $73/105$ ($69.5\%$) human
binary judgments. This comparison is descriptive for the selected,
audit-stratified sample; its class balance differs from the full
Stage~4 population.

\paragraph{Illustrative disagreements and supported exclusions.}
In CASE-066, the auditor excluded a synthetic response expressing
appreciation for \textit{Get Out} based partly on an alleged aversion
to scary films. However, the corresponding historical ReDial user
requested horror and a combination of comedy and scares (ReDial
conversation 3189); two annotators favored retention and one exclusion.
This illustrates the risk of treating genre preferences as exclusive.

Conversely, in CASE-071 all three annotators endorsed the auditor's
exclusion of a detailed recollection from \textit{Arrival}; the
historical user had earlier stated that they did not think they had
seen the film and would add it to a watch list (ReDial conversation
12034). Such temporal evidence is suggestive rather than conclusive:
a user might watch the film subsequently.

Finally, in CASE-082, all three annotators favored excluding a
synthetic response that the auditor left unflagged. The historical
user had previously stated that they had not seen
\textit{The Greatest Showman} (ReDial conversation 14457).
However, expressing a positive opinion about a film does not
necessarily imply having watched it. We therefore treat this case
as a disagreement requiring further examination, rather than
a confirmed auditor error. These examples illustrate turn-level
judgments, not independent validation of individual audit flags.

\paragraph{Scope and limitations.}
The equal $35/35/35$ allocation, together with at most one sampled
turn per historical ReDial user, means that the reported sample
accuracies are not unbiased estimates of accuracy over all Stage~4
turns. Consecutive presentation blocks may confound audit stratum
with order, learning, or fatigue effects. The ReDial excerpts were
retrieved automatically; although matched to historical users, their
relevance and completeness were not exhaustively adjudicated, and
missing context may affect individual decisions. In addition, the
annotators were co-authors and majority vote is an operational
human reference rather than an independently established ground truth.
The evaluation therefore characterizes human agreement with Stage~4
\emph{turn-level} outcomes, not the validity of each issue flag or
population-level recall. The $14$ three-class disagreements,
particularly the five affecting the remove-versus-keep action,
remain priorities for further evidence-based adjudication.
\section{Six Enriched Attribute Schemas}
\label{sec:appendix-enrich}

This appendix provides the formal schema for each enriched attribute 
and an example enriched profile.

\subsection{Schema Definitions}

\paragraph{Aspect-level preferences~\citep{radlinski-etal-2019-coached}.} 
A nested object with three sub-fields. Each entry is an 
\texttt{(anchor, text)} pair where 
\texttt{anchor} $\in$ \{Movie, Genre, Person, Other\}:
\begin{itemize}
\item \texttt{liked}: aspects/items the user expressed positive 
sentiment toward
\item \texttt{disliked}: aspects/items the user expressed negative 
sentiment toward
\item \texttt{descriptions}: factual or descriptive statements 
the user made about specific items
\end{itemize}

\paragraph{Pickiness~\citep{yoon2024evaluating}.} Deterministically
computed from the rejection rate $r$, not LLM-extracted:
\texttt{pickiness\_score} $=5$ if $r\geq0.40$, $4$ if $r\geq0.25$,
$3$ if $r\geq0.10$, $2$ if $r\geq0.05$, and $1$ otherwise.

\paragraph{Decision-making 
style~\citep{Scott1995DecisionMakingST,hayati-etal-2020-inspired}.} 
Object: \texttt{\{style, evidence\}}, where \texttt{style} $\in$ 
\{rational, intuitive, dependent, avoidant, spontaneous\}.

\paragraph{Viewing motivation~\citep{oliver2011entertainment}.} 
Object: \texttt{\{hedonic\_score, primary\_motivation, evidence\}}, 
where \texttt{hedonic\_score} $\in [1, 5]$ ($1$ = purely eudaimonic, 
$5$ = purely hedonic) and \texttt{primary\_motivation} $\in$ 
\{hedonic, eudaimonic, escapism, social, catharsis, arousal\}.

\paragraph{Novelty preference~\citep{tang2022preference}.} Object: 
\texttt{\{novelty\_score, evidence\}}, where 
\texttt{novelty\_score} $\in [1, 5]$.

\paragraph{Domain expertise~\citep{kostric2025know}.} Object: 
\texttt{\{level, evidence\}}, where \texttt{level} $\in$ 
\{novice, intermediate, cinephile\}.

\subsection{Example Enriched Profile (User 468)}

User 468's enriched profile after Stage~5 (interview decision
\textsc{accept}), with values shortened for readability.

\begin{quote}\small\ttfamily
\{
  user\_id: 468,
  aspect\_preferences: \{
    liked: [
      \{Movie: ``John Wick''\},
      \{Genre: ``crime thriller''\}, \{Genre: ``action movie''\},
      \{Other: ``non-stop action and style of choreography''\},
      \{Other: ``high-energy pace and relentless energy''\},
      \{Other: ``physical set pieces and choreography unfold''\}
    ],
    disliked: [],
    descriptions: [
      \{Movie: ``Crouching Tiger, Hidden Dragon rooftop chase lays
                 out history and skill through movement''\},
      \{Other: ``action can do storytelling through style and
                 movement''\}, [...]
    ]
  \},
  pickiness\_score: 1,  // from rejection rate
  decision\_making\_style: \{
    style: ``rational'',
    evidence: ``the steady momentum of a good crime thriller or
               action movie where the pacing never really lets up''
  \},
  viewing\_motivation: \{
    hedonic\_score: 4, primary\_motivation: ``arousal'',
    evidence: ``relentless energy keeps you hooked from the start
               without any wasted moments''
  \},
  novelty\_preference: \{ novelty\_score: 4, [...] \},
  domain\_expertise: \{
    level: ``intermediate'',
    evidence: ``the action itself is doing the storytelling through
               the style and movement''
  \}
\}
\end{quote}

The basic profile that conditions \textsc{StatisticalOnly} contains only 
the Stage 1--2 fields (liked/disliked movie lists, acceptance rate, 
reference style, genre distribution, decade, exploration ratio, 
popularity, and pickiness score). The six enriched attributes above 
are absent in \textsc{StatisticalOnly}: \texttt{aspect\_preferences}, 
\texttt{decision\_making\_style}, \texttt{viewing\_motivation}, 
\texttt{novelty\_preference}, and \texttt{domain\_expertise} are 
all \texttt{null}.

\section{Example End-to-End Simulated Conversations}
\label{sec:appendix-conversations}

This appendix shows conversations between the simulator and the
GPT-4o-mini CRS for the same user (468) under \textsc{Full} and
\textsc{StatisticalOnly}, both with the \textit{mood} opening strategy.

\subsection{\textsc{Full} (opening = mood)}
\begin{quote}\small
\textbf{User:} I'm chasing a real adrenaline rush tonight, so I need an
action thriller with relentless pacing and some genuinely tight
choreography.

\textbf{CRS:} You might enjoy ``John Wick (2014)'' [...]. If you're
looking for something a bit different, ``The Raid: Redemption (2011)''
delivers incredible martial arts and relentless pacing.

\textbf{User:} I've already seen John Wick, but The Raid: Redemption
sounds right up my alley with that martial arts choreography and
pacing. Let's go with that one.

[\textit{\ldots}]

\textbf{User:} That was fantastic, exactly the kind of relentless
pacing and fight choreography I was looking for. Got anything else
along those lines with high energy?
\end{quote}

\subsection{\textsc{StatisticalOnly} (same user, opening = mood)}
\begin{quote}\small
\textbf{User:} I'm in the mood for some intense action tonight.
Something with a good crime or thriller plot.

\textbf{CRS:} How about ``Heat (1995)''? [...]

\textbf{User:} Heat sounds good. I'll watch that one.

[\textit{\ldots}]

\textbf{User:} Die Hard with a Vengeance is a good one. I like those
kinds of action films.
\end{quote}

\subsection{Observations}
The \textsc{Full} simulator voices the aspects recorded in its
enriched profile (pacing, choreography), which the T3 judge extracts
as distinct aspects; the \textsc{StatisticalOnly} simulator refers
mainly to genres and titles. Across this user's four conversations,
T3 retains eight distinct aspects for \textsc{Full} (e.g.,
\textit{pacing}, \textit{martial arts choreography},
\textit{intensity}; filtered entropy $1.75$) but only two for
\textsc{StatisticalOnly} (\textit{plot} and a genre label; entropy
$0.69$), where $11$ of $13$ extracted aspects were generic and
filtered out. The same profile vocabulary also explains the T4
pattern: ``relentless'' appears in three of the four \textsc{Full}
opening requests.

\section{Full System Prompts}
\label{sec:appendix-prompts}

This appendix reproduces the prompts of all LLM-based components in
condensed form. Wording is kept verbatim where shown; ``[...]'' marks
omitted examples or formatting instructions (full versions are in the
released codebase). Text in braces (e.g., \texttt{\{liked\_movies\}})
denotes slots filled at run time.

\subsection{Interviewer Agent (Stage 3)}
\label{sec:appendix-prompts-interviewer}

Each interview has ten messages: an \emph{Account} phase (six
messages) followed by a \emph{Clarify} phase (four messages).
The first interviewer message is not generated: it is one of four
predefined open-ended questions, selected uniformly at random:
\begin{enumerate}
\item ``Tell me about your relationship with movies --- when you sit
down to watch something, what kind of experience are you usually
looking for?''
\item ``Walk me through the last movie you really enjoyed --- what was
it and what stuck with you?''
\item ``If a friend asked you to recommend a movie that captures your
taste, what would you pick and why?''
\item ``When you're choosing what to watch on a typical Friday night,
what goes through your mind?''
\end{enumerate}
All subsequent interviewer messages are generated (temperature $0.7$)
from the interview so far only; the interviewer never receives the
user's profile or ReDial history.

\textbf{Account phase.}
\begin{quote}\small\ttfamily
You are an interviewer conducting a conversation about movie
preferences. Your goal is to understand how this person thinks about
and chooses movies --- their taste, their reasoning, the patterns in
what they like.\\[2pt]
CRITICAL CONSTRAINTS:\\
- You do NOT know anything about this person beforehand. You have no
profile, no history, no list of their favorite movies.\\
- You must NOT make assumptions about what kind of movies they like
based on stereotypes or popular taste patterns.\\
- Every question should follow naturally from what they have ACTUALLY
told you in this conversation.\\[2pt]
QUESTIONING STYLE FOR THIS STAGE (ACCOUNT):\\
- This is the OPENING phase. Let them talk freely.\\
- Ask broad, inviting questions that let them set the direction.\\
- When they mention a specific movie, genre, or feeling, follow up on
THAT thing rather than pivoting to a new topic.\\
- Do NOT challenge or contradict. Avoid leading questions.\\
- Do NOT introduce movies, directors, or genres they did not
mention.\\[2pt]
CONVERSATION SO FAR: \{conversation\_history\}\\
Output ONLY your next question.
\end{quote}

\textbf{Clarify phase.}
\begin{quote}\small\ttfamily
You are continuing the interview from earlier. [...] Your job now is
different: you are NOT collecting more topics --- you are pushing for
the WHY behind what they have already said.\\[2pt]
PREVIOUS DIALOGUE (Stage 1): \{stage\_1\_dialogue\}\\[2pt]
WHAT TO ASK IN THIS STAGE:\\
Pick ONE or TWO things the user said that have a specific texture ---
a particular movie title, a phrase they repeated, a feeling they
named. For each, ask a question that pushes for one of these:\\
- The MECHANISM. They said they liked X --- what specifically about X
works for them?\\
- A CONTRAST. They expressed a general preference --- when does that
preference NOT hold?\\
- The PERSONAL MEANING. They used a word or phrase --- what does it
mean to them concretely?\\[2pt]
WHAT NOT TO DO:\\
- Do NOT ask the same kind of question Stage 1 already asked.\\
- Do NOT zoom out to abstractions.\\
- Do NOT introduce movies, directors, or genres they did not mention.\\
- Do NOT challenge or contradict.\\
- Use THEIR specific vocabulary when probing.\\
{[}... good and bad probe examples ...{]}\\[2pt]
CONVERSATION SO FAR: \{full\_conversation\_history\}\\
Output ONLY your next question.
\end{quote}

\subsection{Interviewee Agent (Stage 3)}
\label{sec:appendix-prompts-interviewee}

The interviewee (temperature $0.6$) receives the user's statistical
profile and one historical ReDial conversation in which the user acted
as a seeker.

\begin{quote}\small\ttfamily
You are participating in a conversation about your movie preferences.
Someone is interviewing you to understand your taste in movies.
Respond naturally, as yourself.\\[2pt]
STATISTICAL PROFILE:\\
- Rejection rate: \{rejection\_rate\} (you accept about
\{accept\_pct\}\% of recommendations on average)\\
- Movies you've liked: \{liked\_movies\}\\
- Movies you've disliked: \{disliked\_movies\}\\
- You tend to talk about movies in terms of: \{dominant\_ref\_style\}\\
- You prefer movies of popularity level: \{popularity\_score\}\\
- Your genre preferences: \{genre\_preferences\}\\
YOUR TASTE PATTERN: \{taste\_summary\}\\
YOUR ACTUAL CONVERSATION HISTORY: \{redial\_conversation\}\\[2pt]
GROUNDING RULES (most important):\\
1. Everything you say must be CONSISTENT with the profile and history
above.\\
2. When you mention specific movies, prefer movies from your
liked/disliked lists or movies that came up in your past
conversation.\\
3. When you give reasoning, the reasoning should be consistent with
your profile patterns.\\
4. Do NOT invent specific personal stories [...] that contradict or go
beyond your profile.\\
5. It is OK and encouraged to say ``I'm not sure'' [...] when a
question pushes beyond what your profile and history can plausibly
answer.\\
6. Do NOT recite your statistical profile out loud.\\[2pt]
SPECIFICITY RULES:\\
- Avoid generic emotional phrases like ``emotional connection'' [...].\\
- When asked WHY you like a film, name a specific element: a scene, a
line, a moment, a character choice, a visual style.\\
- Anchor comparisons in concrete examples from YOUR liked or disliked
lists rather than abstract categories.\\[2pt]
CONVERSATION STYLE: first person; 2--4 sentences; avoid absolute
statements unless your profile genuinely supports them. [...]\\
CONVERSATION SO FAR: \{conversation\_history\}
\end{quote}

When the audit decision is \textsc{regenerate}, the interview is
regenerated (up to two retries) with the following nudge appended to
the interviewee prompt:

\begin{quote}\small\ttfamily
ADDITIONAL GUIDANCE FROM A PREVIOUS ATTEMPT:\\
In your previous attempt at this conversation, you made claims that
were not fully grounded in your profile or history. Be particularly
careful to:\\
- avoid absolute statements unless your profile supports them\\
- avoid inventing specific personal anecdotes\\
- prefer ``I'm not sure'' over fabricated detail
\end{quote}

\subsection{Audit Agent (Stage 4)}
\label{sec:appendix-prompts-auditor}

The Audit Agent combines a deterministic structural check with three
separate LLM calls (Claude Haiku 4.5, temperature $0.2$). Only
interviewee turns are audited.

\textbf{(a) Structural check (deterministic).} Verifies that the
interview has ten turns, that interviewer and interviewee strictly
alternate, and that no turn is empty. Any violation yields a
high-severity \texttt{structural\_error}, and the LLM checks are
skipped.

\textbf{(b) Movie mentions vs.\ profile.}
\begin{quote}\small\ttfamily
Extract all movie titles mentioned by the USER (not the interviewer)
in this dialogue. For each movie, determine the sentiment expressed
about it. [...] Classify sentiment: ``positive'', ``negative'', or
``neutral''.
\end{quote}
A movie mentioned positively but listed as disliked in the profile, or
negatively but listed as liked, is flagged deterministically as a
high-severity \texttt{history\_contradiction}.

\textbf{(c) Contradiction with the ReDial conversation.}
\begin{quote}\small\ttfamily
You are checking whether a recent dialogue contradicts an earlier
conversation by the same user.\\
EARLIER CONVERSATION: \{redial\_conversation\}\\
RECENT DIALOGUE TO CHECK: \{dialogue\_text\}\\
TASK: For each statement made by the USER [...], check whether it
directly contradicts something they said in the earlier conversation.
Distinguish carefully:\\
- CONTRADICTION: directly opposing statements\\
- REPHRASING / ELABORATION / UNRELATED: NOT a contradiction\\
Only flag CONTRADICTION. Output: turn\_index, user\_statement,
redial\_reference, explanation, confidence (low|medium|high).
\end{quote}
Each contradiction becomes a \texttt{history\_contradiction} flag whose
severity equals the reported confidence.

\textbf{(d) Fabrication and caricature.}
\begin{quote}\small\ttfamily
You are auditing a dialogue where a user discussed their movie
preferences. The user was supposed to respond consistently with the
profile and prior conversation history below. [...]\\[2pt]
CATEGORY 1 --- FABRICATION\\
A specific factual claim that is neither in the profile nor plausibly
inferable from it. The bar for flagging is HIGH --- only flag specific,
ungrounded claims.\\
DO flag: personal anecdotes with specific people, places, or dates;
claims about working in the film industry, attending specific
festivals, knowing directors; specific verifiable facts the user could
not plausibly know.\\
DO NOT flag: plausible elaboration on a movie in the profile; general
emotional reactions; vague time references; hedged statements.\\[2pt]
CATEGORY 2 --- CARICATURE\\
A response that performs a stereotype rather than reflecting the
actual profile. The test: would someone with THIS specific profile
plausibly say this, or is this what an LLM imagines a ``type of
person'' would say?\\
DO flag: absolute statements that contradict profile nuance; reciting
tropes about a ``type of person''; extreme rejection or acceptance
beyond what the rejection rate supports; a snob/cinephile persona
without profile support.\\
DO NOT flag: proportionate strong opinions; specific personal
preferences; self-aware statements; mixed feelings.\\[2pt]
Output: for each issue, turn\_index, claim or quote, reason, and
severity (low|medium|high).
\end{quote}

\textbf{Decision rule (deterministic).} Any \texttt{structural\_error}
or $\geq 3$ high-severity flags yield \textsc{regenerate}; at least one
high-severity or $\geq 3$ medium-severity flags yield \textsc{partial},
in which case all high- and medium-severity interviewee turns are
removed from the enrichment input; otherwise \textsc{accept}.

\subsection{Profile Enrichment (Stage 5)}
\label{sec:appendix-prompts-enrich}

A single Tencent Hy3 call (temperature $0.1$) receives the audited
interview and the user's Stage~2 profile as background context.

\begin{quote}\small\ttfamily
You are an expert in user modeling for movie recommender systems.
[...]\\
You are given two inputs: 1. The user's existing profile data [...]
2. A movie interview dialogue with that same user. Use the profile
data as background context to better interpret what the user says in
the dialogue. Extract exactly 6 profile fields based on what the user
says in the dialogue.\\
USER PROFILE (background context): \{profile\_context\}\\
INTERVIEW DIALOGUE: \{dialogue\_text\}\\[2pt]
FIELD DEFINITIONS:\\
1. aspect\_preferences: items of \{anchor: Movie|Genre|Person|Other,
text\}, grouped into liked, disliked, and descriptions (neutral
observations). Extract only from what the user says.\\
2. pickiness: DO NOT EXTRACT from dialogue [...]. Set to null.\\
3. decision\_making\_style: \{style:
rational|intuitive|dependent|avoidant|spontaneous, evidence\}\\
4. viewing\_motivation: \{hedonic\_score: 1--5 (eudaimonic
$\leftrightarrow$ hedonic), primary\_motivation:
hedonic|eudaimonic|escapism|social|catharsis|arousal, evidence\}\\
5. novelty\_preference: \{novelty\_score: 1--5 (conservative
$\leftrightarrow$ explorer), evidence\}\\
6. domain\_expertise: \{level: novice|intermediate|cinephile,
evidence\}\\
{[}... definitions of each category and score level ...{]}\\[2pt]
RULES:\\
- Extract only from the DIALOGUE, not from the profile context\\
- Use the profile context only to resolve ambiguity or calibrate
scores\\
- If a field cannot be determined from the dialogue, set it to null\\
- Do not invent information not present in the dialogue\\
Return ONLY this JSON [...]
\end{quote}

The pickiness score ($1$--$5$) is assigned deterministically from the
user's rejection rate $r$: $5$ if $r\geq0.40$, $4$ if $r\geq0.25$, $3$
if $r\geq0.10$, $2$ if $r\geq0.05$, and $1$ otherwise.

\subsection{User Simulator (Stage 6)}
\label{sec:appendix-prompts-simulator}

\textbf{Opening message.} For each of the four opening strategies, the
simulator first generates the opening request (temperature $0.7$):
\begin{quote}\small\ttfamily
You are generating the OPENING MESSAGE for a user about to start a
conversation with a movie recommender.\\
USER PROFILE: \{profile\_context\}\\
OPENING STRATEGY: \{strategy\}\\
Generate ONE opening message (1--3 sentences) that:\\
- Uses the strategy specified above\\
- Reflects this user's decision style (\{decision\_style\}) and
expertise level (\{expertise\_level\})\\
- Is specific enough to give the recommender something to work with
[...]\\
- Sounds natural, like a real person typing\\[2pt]
Strategies: aspect (signal one or two specific aspects from your
liked-aspects list); movie (anchor on a movie from your liked list and
ask for something similar); mood (describe the mood you want the movie
to match tonight); expertise (signal domain knowledge before asking
for something specific). [... example phrasings per strategy; do not
copy any example verbatim ...]
\end{quote}

\textbf{Simulator.} Subsequent user turns (temperature $0.7$) are
generated with:
\begin{quote}\small\ttfamily
You are simulating a real person seeking movie recommendations from a
recommender system.\\
\{profile\_context\}\\[2pt]
CRITICAL BEHAVIORAL RULES:\\
1. You are the USER. NEVER take the role of the recommender. [...]\\
2. Do NOT prefix your messages with any role label.\\
3. Do NOT recite your profile aloud. Your preferences emerge through
your reactions, not declarations.\\
4. Stay consistent with your profile across all turns.\\
5. Keep responses 1--3 sentences typically.\\
6. \{pickiness\_rule\_block\}\\
7. Do not invent new movies you've watched beyond your profile.
\end{quote}
The pickiness block depends on the pickiness score: \emph{agreeable}
for scores $\leq 2$ (``Default to accepting recommendations unless
they obviously contradict your stated preferences''), \emph{moderate}
for $3$ (``Accept recommendations that match at least TWO of your
stated preferences''), and \emph{picky} for $\geq 4$ (``Reject
recommendations that fail to match your stated preferences in
specific ways [...] state the specific reason''). Under
\textsc{StatisticalOnly}, \texttt{\{profile\_context\}} omits the
enriched attributes.

\textbf{CRS (GPT-4o-mini, temperature $0.5$).}
\begin{quote}\small\ttfamily
You are a conversational movie recommender system. [...] Ask
clarifying questions when the user's request is vague; recommend 1--3
specific movies with brief reasoning when you have enough information;
provide titles in the format ``Title (Year)''; adapt to user feedback;
keep responses concise (2--4 sentences typically).
\end{quote}

After each CRS turn, starting from the fourth, the termination judge
(\S\ref{sec:appendix-prompts-eval}) classifies the last four messages;
any label other than \textsc{continue} ends the conversation, which is
otherwise capped at eight CRS turns.

\subsection{Evaluation Judges}
\label{sec:appendix-prompts-eval}

In the primary experiment, all judges use Claude Haiku 4.5 at
temperature $0.0$. The termination judge operates during simulation;
the others operate on the stored conversations. In the separate
cross-CRS experiment, GPT-4o-mini serves as the termination judge and
performs the post-hoc conversation audit before T3 extraction, without
access to the extraction outputs; separate GPT-4o-mini and
DeepSeek-chat calls then evaluate T3
(Appendix~\ref{sec:appendix-cross-crs}).

\textbf{Termination judge (Stage 6).}
\begin{quote}\small\ttfamily
Classify whether the following conversation has reached a natural
endpoint. [...] Output EXACTLY ONE of these labels:\\
- true\_acceptance: user explicitly endorses a SPECIFIC recommendation
with positive language\\
- resigned\_acceptance: user acknowledges a recommendation without
enthusiasm or specificity\\
- explicit\_giveup: user explicitly disengages or expresses
frustration\\
- other: conversation has ended but fits none of the above\\
- CONTINUE: user is still exploring or asking\\
RULES: true\_acceptance requires BOTH a specific title AND clearly
positive language; do NOT output it if the user asks a follow-up
question.\\
CONVERSATION (last 4 messages): \{last\_messages\}
\end{quote}

\textbf{Movie mention extraction (T1, T2, T3).}
\begin{quote}\small\ttfamily
Identify every distinct movie that appears in the conversation. For
each one, return: title (canonical); year; introduced\_by (``user'' if
the USER first mentioned this movie, ``crs'' if the CRS introduced
it); first\_user\_mention\_turn\_index; all\_turn\_indices. Do NOT
include person names, countries, or generic words. Do NOT include
hypothetical references like ``a comedy''. Deduplicate surface forms.
[...]
\end{quote}

\textbf{Rating inference (T2).}
\begin{quote}\small\ttfamily
Infer how much the user would rate a specific movie. Assign a rating
ONLY when the user expresses an actual opinion or evaluative reaction
about this movie itself. REFERENCE-ONLY RULE: if the user mentions
this movie ONLY as a reference point or comparison anchor (e.g.,
``something like X''), the rating MUST be null. Scale: 5.0 loved /
accepted enthusiastically; 4.0 clear positive interest; 3.0 neutral;
2.0 mild disinterest; 1.0 explicitly rejected (half points allowed).
Output: \{rating, confidence: high|medium|low, evidence\}.
\end{quote}

\textbf{Aspect--sentiment extraction (T3).}
\begin{quote}\small\ttfamily
You are extracting aspect-sentiment pairs from a user's opinion about
a movie. [...] Extract ALL aspect-sentiment pairs the user mentions.
Aspects are concrete elements of the movie: plot, acting, pacing,
cast, soundtrack, cinematography, dialogue, ending, characters, mood,
atmosphere, etc. For each pair output: aspect (short noun phrase);
sentiment (positive|neutral|negative); evidence (short quote from the
user). If user expressed no specific aspect opinions, output no pairs.
\end{quote}
A post-processing step removes generic placeholders
(\textit{movie(s)}, \textit{film(s)}, \textit{overall movie},
\textit{overall film}, \textit{story/stories}, \textit{thing(s)},
\textit{stuff}, \textit{experience}) before computing aspect entropy.

\textbf{Acceptance/rejection classification.}
\begin{quote}\small\ttfamily
For each movie the CRS explicitly recommended, classify how the USER
responded to it. ONLY classify movies the CRS recommended.\\
- accept: clear positive intent to watch, enthusiasm, or agreement to
try it (``sounds great'', ``I'll watch that'')\\
- reject: explicit decline, clear disinterest, or dismissal (``not
really my thing'', ``no thanks'')\\
- neutral: ambiguous, hedging, or absent reaction (``maybe'', no
direct response)
\end{quote}

\section{Model Configuration and API Cost}
\label{sec:appendix-config}

Table~\ref{tab:appendix-models} lists the model, decoding temperature,
and output-token limit of every LLM component in the primary
experiment. Reasoning was disabled for Gemini 2.5 Flash and enabled
(effort \textit{high}) for Tencent Hy3; all other models were used
without reasoning.

\begin{table}[t]
\centering
\footnotesize
\setlength{\tabcolsep}{3pt}
\begin{tabular}{@{}llrr@{}}
\toprule
\textbf{Component} & \textbf{Model} & \textbf{Temp.} & \textbf{Max tok.} \\
\midrule
Interviewer (S3)          & Gemini 2.5 Flash & 0.7 & 512 \\
Interviewee (S3)          & Gemini 2.5 Flash & 0.6 & 512 \\
Audit: movie check (S4)   & Claude Haiku 4.5 & 0.2 & 512 \\
Audit: contradiction (S4) & Claude Haiku 4.5 & 0.2 & 512 \\
Audit: fabr./caric.\ (S4) & Claude Haiku 4.5 & 0.2 & 1{,}024 \\
Enrichment (S5)           & Tencent Hy3      & 0.1 & 8{,}192 \\
Opening message (S6)      & Gemini 2.5 Flash & 0.7 & 256 \\
User simulator (S6)       & Gemini 2.5 Flash & 0.7 & 256 \\
CRS (S6)                  & GPT-4o-mini      & 0.5 & 256 \\
Termination judge (S6)    & Claude Haiku 4.5 & 0.0 & 20 \\
\midrule
Movie extraction (T1--T3) & Claude Haiku 4.5 & 0.0 & 2{,}000 \\
Rating inference (T2)     & Claude Haiku 4.5 & 0.0 & 256 \\
Aspect--sentiment (T3)    & Claude Haiku 4.5 & 0.0 & 2{,}048 \\
Accept/reject             & Claude Haiku 4.5 & 0.0 & 1{,}000 \\
\midrule
CSHI simulator            & Gemini 2.5 Flash & 0.0 & 2{,}048 \\
\bottomrule
\end{tabular}
\caption{LLM components of the primary experiment. S3--S6 denote
pipeline stages. The CSHI baseline uses the same CRS configuration.}
\label{tab:appendix-models}
\end{table}
\paragraph{API cost.}
Table~\ref{tab:appendix-cost} reports the provider-reported API cost
of each run, split into generation (Stages~3--6, including the CRS)
and evaluation (all judges). Generating the full \textsc{Full} cohort
cost \$8.30, i.e., about \$0.028 per user including four simulated
conversations; profile enrichment alone (Stages~3--5) cost \$4.29
(\$0.014 per user). Evaluation dominates the total cost, chiefly
through per-movie rating inference for T2. \textsc{StatisticalOnly}
and CSHI involve no interviews or enrichment, so their generation
cost covers only user simulation and the CRS.

\begin{table}[t]
\centering
\footnotesize
\setlength{\tabcolsep}{3.5pt}
\begin{tabular}{@{}lrrrrr@{}}
\toprule
\textbf{Run} & \textbf{Calls} & \textbf{Tokens} & \textbf{Gen.} & \textbf{Eval.} & \textbf{Total} \\
 & & \textbf{in/out (M)} & \textbf{(\$)} & \textbf{(\$)} & \textbf{(\$)} \\
\midrule
\textsc{Full} (300)     & 32{,}907 & 28.7 / 4.4 & 8.30 & 23.85 & 32.15 \\
\textsc{StatOnly} (100) & 11{,}149 &  8.5 / 0.9 & 1.41 &  8.93 & 10.34 \\
CSHI (100)              & 16{,}589 & 14.8 / 1.4 & 1.37 & 16.90 & 18.27 \\
\midrule
\textbf{Total}          & \textbf{60{,}645} & \textbf{52.1 / 6.8} & \textbf{11.08} & \textbf{49.68} & \textbf{60.76} \\
\bottomrule
\end{tabular}
\caption{API usage and cost (USD) per run of the primary experiment.
Hy3 output tokens include reasoning tokens. Numbers in parentheses
give the number of users.}
\label{tab:appendix-cost}
\end{table}

\end{document}